%% file: main.tex
\pdfoutput=1
\documentclass[11pt]{article}

\usepackage[final]{acl}

\usepackage{times}
\usepackage{latexsym}
\usepackage[T1]{fontenc}
\usepackage[utf8]{inputenc}
\usepackage{microtype}
\usepackage{inconsolata}
\usepackage{graphicx}
\usepackage{booktabs}
\usepackage{multirow}
\usepackage{amsmath}
\usepackage{amssymb}
\usepackage{algorithm}
\usepackage{algpseudocode}
\usepackage{tcolorbox}
\tcbuselibrary{listings,breakable}
\newtcblisting{promptbox}{
  listing only, breakable, colback=gray!4, colframe=gray!60, boxrule=0.5pt, arc=1mm,
  left=1.5mm, right=1.5mm, top=1mm, bottom=1mm,
  listing options={basicstyle=\ttfamily\footnotesize, breaklines=true, breakatwhitespace=false,
    columns=fullflexible, keepspaces=true, postbreak=\mbox{{\footnotesize$\hookrightarrow$}\,},
    literate={—}{{---}}1 {≥}{{$\geq$}}1}}

\title{Evidence Attribution in Visual Document Understanding \\ without Coordinates or Region Labels}

\author{Zhuchenyang Liu, Yao Zhang, Yu Xiao \\
  Aalto University \\
  Espoo, Finland \\
  \texttt{zhuchenyang.liu@aalto.fi}}

\begin{document}
\maketitle

\begin{abstract}
Reliable visual document understanding requires a model to attribute each answer to the
evidence regions that support it. Recent benchmarks and systems express this step through a
coordinate interface: the model outputs the coordinates of bounding boxes that mark the
evidence regions in the document. Under this interface, vision-language models often fail to
identify the right regions even when the answer is correct, a failure known as Attribution
Hallucination. We present a study that investigates whether this failure is
partially limited by what the model can express through coordinates. On a verified bilingual
CiteVQA subset, we compare the coordinate interface with a language interface in which the
model outputs only text, quoting its evidence verbatim, and a multimodal retriever returns
the location of each quote as a page region proposed by a layout parser (tables and figures
are quoted through their captions or notes); the comparison is repeated over six open
vision-language models. Compared with the coordinate interface, evidence recall rises from at
most 8 points to between 26 and 47 and the hallucination rate roughly halves, with little
change in answer quality.
Building on this comparison, we use the same
quote-and-retrieve pipeline as a training scaffold: because region-level evidence labels are
expensive to collect for long documents, we introduce a GRPO recipe whose reward is a judge's
reading of the gold answer and crops of the retrieved regions, training the model to quote
better evidence without any region labels and raising an 8B backbone's strict attributed
accuracy from 22.4 to 33.8. These findings indicate a practical path to improve attribution without a coordinate interface and without costly region-level supervision.\footnotemark
\end{abstract}

\section{Introduction}

\begin{figure}[h]
\centering
\includegraphics[width=1\columnwidth]{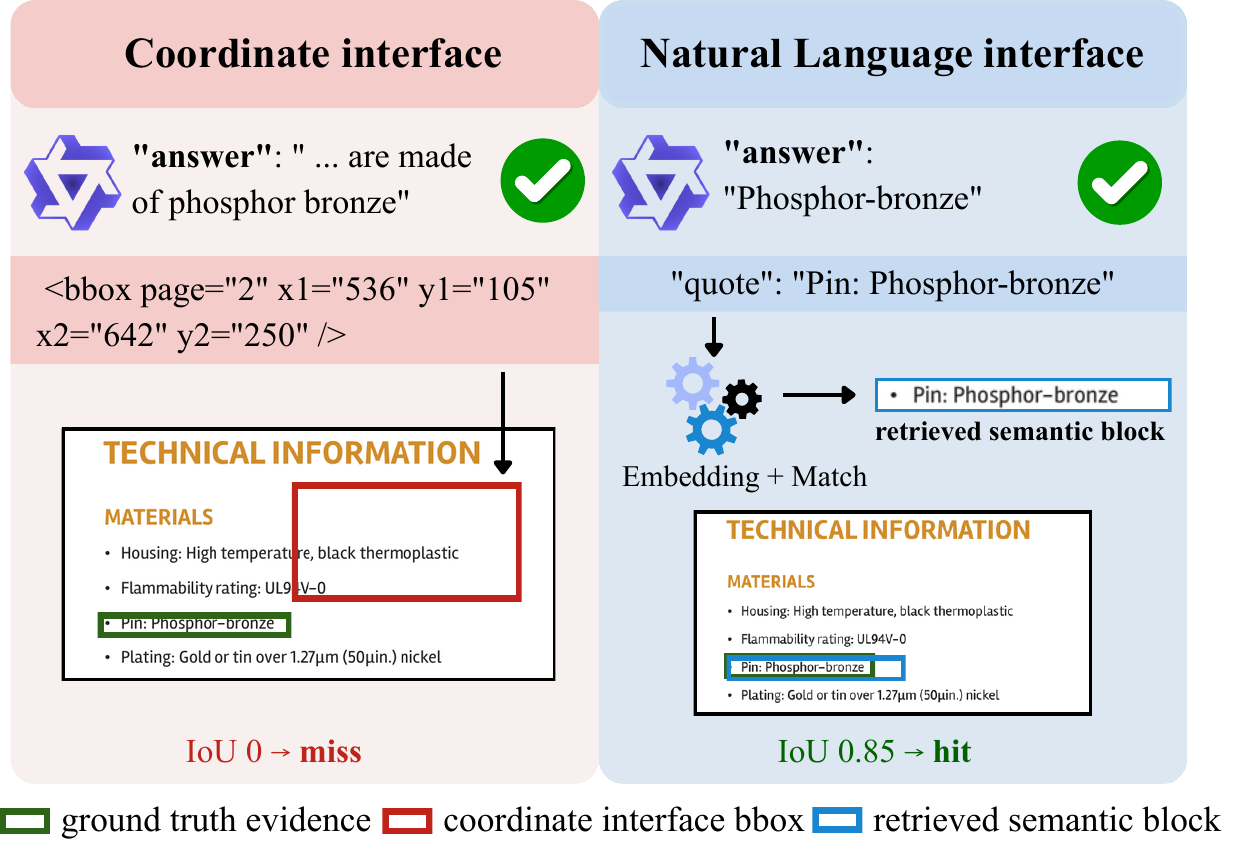}
\caption{The interface artifact in one example (Qwen3-VL-8B; page 2 of a 24-page input;
question: what specific alloy is used for the internal contact pins). Both answers are
correct. Only the evidence interface differs.}
\label{fig:teaser}
\end{figure}

As vision-language models (VLMs) move from single pages to long, layout-rich documents, visual
document understanding demands more than a correct answer: in domains such as finance, law,
and medicine, an answer is useful only if it can be verified, and verification requires
evidence attribution, the model pointing to the regions that support each answer.
This requirement has recently crystallized into a line of benchmarks and systems: CiteVQA
\citep{ma2026citevqabenchmarkingevidenceattribution} scores answers jointly with element-level evidence citations, where
an element is a single paragraph, table, or figure on a page; further benchmarks annotate
bounding-box evidence \citep{loison-etal-2026-vidore, yu2026sciegqadatasetscientificevidencegrounded}; and attribution systems
train or prompt models to localize their evidence
\citep{ma-etal-2025-visa, 10.1609/aaai.v40i38.40488, Liu_2026}. Across all of these, evidence is
expressed through the same interface: the model outputs the coordinates of a bounding box.

\footnotetext{Code, trained model, and data: \url{https://github.com/Ryenhails/quote-and-retrieve}}%
Under this coordinate interface, a model attributes evidence by autoregressively generating
numeric coordinate tokens, a format inherited from object detection. The reported results are
discouraging. Auditing twenty models, CiteVQA finds that even strong systems frequently produce
a correct answer while citing a wrong region, a failure it names Attribution Hallucination, and
open models rarely exceed single-digit evidence recall. CiteVQA reads this failure as a
lack of fundamental grounding reliability.

This reading raises two issues, and we address them in turn. The first is diagnostic: is the
failure a missing capability, or an artifact of the coordinate interface? Coordinate-based
document grounding was posed only recently \citep{ma2026citevqabenchmarkingevidenceattribution, loison-etal-2026-vidore, yu2026sciegqadatasetscientificevidencegrounded}, so VLM training may offer little direct supervision for it, and no prior
work compares evidence interfaces on the same models in visual documents. Evidence from other
domains, however, points at the output format itself as the weak link: replacing or bypassing
coordinate emission recovers grounding in interface agents and natural images
\citep{wu2025guiactorcoordinatefreevisualgrounding, yang2023setofmarkpromptingunleashesextraordinary, kang2025largevisionlanguagemodelneeds}, format restrictions degrade
language model performance more broadly \citep{tam-etal-2024-speak}, and object-grounding objectives
do not reliably reduce hallucination \citep{geigle-etal-2024-object}. The apparent lack of
capability may therefore be a mismatch between the evidence a model can express in language
and what the coordinate interface can carry. The second issue is practical: systems that
respond to the failure train on region or page labels \citep{ma-etal-2025-visa, 10.1609/aaai.v40i38.40488, xiong2025docr1evidencepageguidedgrpo}, yet region-level evidence labels are expensive to collect for long
documents, so we ask whether attribution can be improved without them.

The first issue calls for a comparative study between interfaces: if models already express location-relevant information that the coordinate format cannot carry, a more expressive evidence interface should recover it. We hold the backbone, input pages, questions, and scoring fixed and change only the way the
model expresses evidence (Figure~\ref{fig:teaser}). In the coordinate condition, the model
follows the CiteVQA protocol and emits bounding-box coordinates. In the language condition,
the model outputs only text: it quotes its evidence verbatim, a layout parser divides each
page into candidate regions, semantic blocks, and a multimodal retriever returns for each
quote the block it comes from. This condition adapts quote-based citation, an established
practice in text RAG \citep{gao-etal-2023-enabling, zhang-etal-2025-longcite}, to page regions in
visual documents. The interface complements coordinate prediction as a model-agnostic post-processing
layer: the model localizes in language, and the parser and the retriever ground that language
into regions. Our study compares the two evidence interfaces on fixed backbones. We evaluate on the single-document portion of CiteVQA, restricted to 719
bilingual questions whose annotations we could verify against the available source PDFs.

The comparison recovers most of the reported failure. Across six open models from four
families, spanning 8B to 31B parameters, evidence recall rises from at most 8.1 under
coordinates to between 25.9 and 46.9 under the language interface, and the hallucination rate
falls from between 82 and 97 percent to between 39 and 65 percent, with little change in
answer quality. Because the language condition adds the parser and the retriever, we run
ablation controls that apportion the credit; they show that the recovery rests on location
information that the models express in generated language.

For the second issue, the same pipeline becomes a training scaffold that removes the need for
region-level evidence labels. We construct a reward in which a vision-language judge compares
the model answer against the gold answer and scores crops of the cited regions, so the signal
derives entirely from data that any QA set already contains. GRPO turns these judged rewards
into group-relative advantages for the quoting policy, with DAPO's asymmetric clipping for
stability \citep{shao2024deepseekmathpushinglimitsmathematical, yu2025dapoopensourcellmreinforcement}. Trained on 1.6K questions built from
LongDocURL \citep{deng-etal-2025-longdocurl} and instantiated on the 8B backbone, the smallest
with reliable citation formatting, the model reaches 51.3 evidence recall and a 28.4 percent
hallucination rate, and an independent evaluation judge suggests that the improvement
reflects better evidence rather than reward exploitation.

Our contributions are threefold. First, a controlled study that, to our knowledge, is the
first to analyze the impact of the evidence interface in visual document attribution,
quantified through an explicit Attribution Hallucination rate. Second, a language evidence interface that carries
quote-based citation, established practice in text RAG, across the modality boundary,
resolving quotes to pixel regions. Third, a region-label-free reinforcement learning recipe that raises an 8B
backbone's strict attributed accuracy from 22.4 to 33.8.

\section{Related Work}

Four lines of work border ours: text-QA attribution supplies the quote-then-resolve pattern
we transfer, extractive Doc-VQA delimits our task, coordinate-based evidence systems are the
object of our diagnosis, and reinforcement learning for grounding is the paradigm our reward
modifies.

\paragraph{Evidence attribution in text-based QA.}
Attribution has been studied extensively in settings where both the answer and the evidence are
text. GopherCite trains models with reinforcement learning to support answers with verbatim
quotes \citep{menick2022teachinglanguagemodelssupport}, ALCE formalizes citation evaluation
\citep{gao-etal-2023-enabling}, RARR attributes generations through post-hoc retrieval
\citep{gao-etal-2023-rarr}, and LongCite produces fine-grained span citations from automatically
constructed supervision \citep{zhang-etal-2025-longcite}. Citation quality can be improved without
human labels \citep{cohenwang2024contextciteattributingmodelgeneration, chuang2025selfciteselfsupervisedalignmentcontext} and evaluated for
whether citations support the statements \citep{10.1145/3731120.3744592}. However, our evidence is a region of a rendered page
rather than a text span, so quotes cross a modality boundary and are scored geometrically.

\paragraph{Answer localization in extractive Doc-VQA.}
In extractive document QA, the answer is a text
span that appears verbatim on the page, so a system can return a box around that span as a form
of explanation. DLaVA \citep{mohammadshirazi2025dlavadocumentlanguagevision} tags detected text regions with identifiers so that the
model can name the answer region, and ARIAL \citep{mohammadshirazi2025arialagenticframeworkdocument} maps the generated answer string back
to an OCR segment. In our task setting, most questions are abstractive syntheses over long
documents, therefore what has to be localized is the supporting
evidence, often several elements on different pages.

\paragraph{Localizing evidence in visual documents.}
Closest to our task are systems that localize supporting evidence in rendered documents. VISA
\citep{ma-etal-2025-visa} learns visual source attribution from large-scale box supervision, LAT
\citep{10.1609/aaai.v40i38.40488} combines stepwise box prediction with reinforcement learning, and Chain
of Evidence \citep{Liu_2026} prompts a model to emit pixel-level boxes along its
reasoning chain. All of them express evidence directly as coordinates, and none examines the
interface itself. These systems are trained and evaluated on a few candidate page
screenshots only.

\paragraph{Reinforcement learning for visual grounding.}
Verifiable-reward reinforcement learning has improved visual grounding in general settings,
for detection, referring, and grounded reasoning tasks
\citep{liu2025visualrftvisualreinforcementfinetuning, shen2025vlmr1stablegeneralizabler1style, cao2026groundr1incentivizinggroundedvisual}, with annotation-free variants
pursued through closed-loop frameworks \citep{yang2026harthighresolutionannotationfreereasoning}, and subsequently for multi-page
documents \citep{xiong2025docr1evidencepageguidedgrpo}. These methods compute rewards on predicted coordinates or
page identities and therefore presuppose region or page labels. Our reward is
computed on natural-language quotes after retrieval, requires neither kind of label, and gates
on answer correctness.

\section{Method}
\label{sec:method}

\paragraph{Problem statement.}
A document $D$ is a sequence of page images $(p_1,\dots,p_N)$. Given a question $q$, the system outputs an answer $y$ along with a citation set $C=\{(k_j, b_j)\}_{j=1}^{m}$, where $k_j$ is a page index and $b_j$ is a bounding box on that page. The ground truth provides the answer $y^{*}$ and the necessary evidence elements $G=\{(k^{*}_i, b^{*}_i)\}$. 

We say that a cited box $b$ matches a ground-truth element $b^{*}$ on the same page if their
intersection over union, $\mathrm{IoU}(b, b^{*}) = \lvert b \cap b^{*}\rvert / \lvert b \cup
b^{*}\rvert$, is at least $0.5$; recall and the remaining metrics are defined in
Section~\ref{sec:setup}. We distinguish between two tasks: page-level localization, which identifies the correct page, and element-level localization, which identifies the annotated unit on that page—whether a paragraph, table, or figure. Ground truth is annotated at the element level, and we call the units a layout parser proposes semantic blocks. 

\begin{figure*}[htbp]
\centering
\includegraphics[width=1\textwidth]{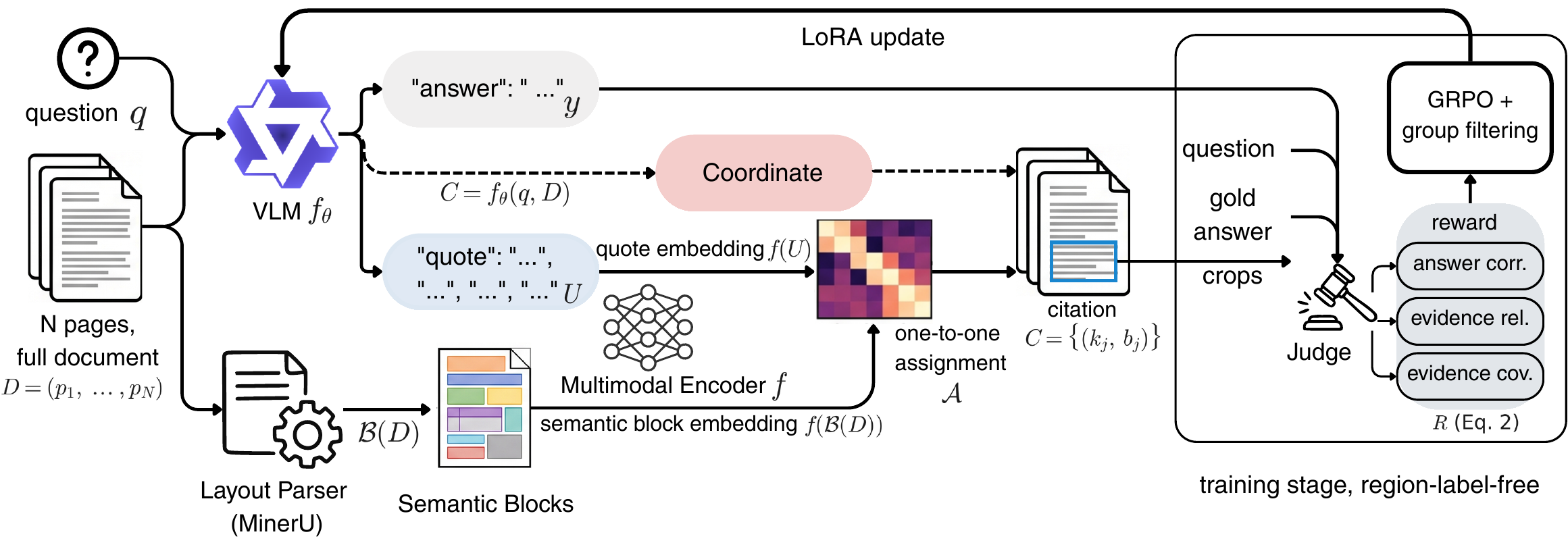}
\caption{Method overview. Solid lines denote our language interface: a layout parser proposes semantic blocks, a multimodal encoder embeds both the model's verbatim quotes and the proposed blocks, and a one-to-one assignment matches each quote to a block, whose page and box become the citation. Dashed lines denote the coordinate interface, which directly outputs box coordinates and is evaluated on those raw outputs. The boxed region on the right illustrates the training phase: a judge compares the retrieved crops and the model answer against the gold answer, and the gated reward from Eq.~\ref{eq:reward} drives GRPO optimization.}
\label{fig:pipeline}
\end{figure*}

The two interfaces differ in how $C$ is produced (Figure~\ref{fig:pipeline}). Under the coordinate interface, the model generates $C$ directly as coordinate tokens: $C = f_\theta(q, D)$. Under our language interface, the model instead produces verbatim quotes $U = f_\theta(q, D)$; the citation set is then assembled outside the model as 
$$
C = \mathcal{A}\big(f(U),\, f(\mathcal{B}(D))\big),
$$
where $f$ embeds the quotes and the parser-proposed blocks $\mathcal{B}(D)$, and $\mathcal{A}$ performs a one-to-one assignment (Section~\ref{sec:retrieval}). Under this interface the model localizes in language and the external components ground that language into regions, so the interface applies to any backbone without modification. This interface further enables training for attribution without requiring region-level labels (Section~\ref{sec:rl}). 

\subsection{Natural-language evidence interface}
\label{sec:interface}
In contrast to coordinate prediction, our model produces evidence in its native textual modality: a JSON object comprising a free-form answer and a list of supporting quotes. Each quote is a verbatim span extracted from a single page region, with spans drawn at the level of a sentence, table row, table cell, caption, or note. This design naturally handles non-textual evidence as well, since tables and figures are represented through their captions, cells, or notes. Requiring verbatim spans also anchors each quote to a specific block: a caption or a cell is nearly unique within a document, whereas a free-form description of a figure could match any topically similar one under semantic retrieval. Moreover, because the answer generation mechanism is identical to standard question answering, answer quality is unlikely to be affected by the interface change. The complete prompt is given in Appendix~\ref{app:prompt-nl}.

\subsection{Whole-document semantic block retrieval}
\label{sec:retrieval}
Quotes are resolved to regions by retrieving from the parsed document structure (Figure~\ref{fig:pipeline}, center). A layout parser (MinerU \citep{wang2024mineruopensourcesolutionprecise}) segments every page into typed semantic blocks, each representing a paragraph, table, figure, or caption. Each block $\beta$ in $\mathcal{B}(D)=\{\beta_1,\dots,\beta_n\}$ carries a page index $k(\beta)$ and a bounding box $b(\beta)$.

To enable retrieval, we render each block as an image crop and embed it with a multimodal encoder $f$ (Qwen3-VL-Embedding-2B \citep{li2026qwen3vlembeddingqwen3vlrerankerunifiedframework}). Embedding the crop rather than the extracted text is crucial, as about thirty percent of evidence is non-textual, and rendered tables often have low similarity to plain-language quotes in the text space. Each quote $u_j$ is embedded with the same encoder, yielding similarities $S_{j\ell} = \cos\!\big(f(u_j),\, f(\beta_\ell)\big)$ for all blocks.

We consider candidates from the entire document and deliberately ignore the model's predicted page numbers, which are unreliable, to prevent page errors from becoming unrecoverable. Quotes are then matched to blocks by solving a linear assignment problem:
\begin{equation}
\label{eq:assign}
\pi^{*} = \arg\max_{\pi \in \Pi}\; \sum_{j=1}^{m} S_{j,\pi(j)},
\end{equation}
where $\Pi$ is the set of injective maps from the $m$ quotes to the $n$ blocks, ensuring that each block receives at most one quote. We solve Eq.~\ref{eq:assign} using the Hungarian algorithm \citep{kuhn1955hungarian}, and the final citation set becomes
$C=\{(k(\beta_{\pi^{*}(j)}),\, b(\beta_{\pi^{*}(j)}))\}_{j=1}^{m}$.

\subsection{Region-label-free reinforcement learning}
\label{sec:rl}
Our training signal does not require region-level labels (Figure~\ref{fig:pipeline}, right). The reward is constructed from the question, model answer, gold answer, and retrieved crops of the page regions the model cites. We instantiate this reward on the language interface, where the base policy cites evidence with sufficient frequency for group-relative optimization to carry meaningful signal, and we optimize using Group Relative Policy Optimization (GRPO; \citealp{shao2024deepseekmathpushinglimitsmathematical}).

For each question, the policy samples $n$ responses. Each response is parsed into an answer and a list of quotes, which are then resolved to block crops via the retrieval procedure described in Section~\ref{sec:retrieval}. A vision-language judge takes as input the question, the gold answer, the model's answer, and the retrieved crops, and outputs three integer scores in $[0,5]$: answer correctness $a$, evidence relevance $e_{\mathrm{rel}}$, and evidence coverage $e_{\mathrm{cov}}$ (prompt in Appendix~\ref{app:prompt-reward}). Importantly, the judge operates on the retrieved crops rather than the raw quotes, so that a fluent but mismatched quote—one assigned to the wrong block—receives a low score, keeping the reward aligned with box-level evaluation metrics.

The reward is defined as
\begin{equation}
\label{eq:reward}
r = \frac{a}{5}\cdot\frac{e_{\mathrm{rel}}+e_{\mathrm{cov}}}{10}.
\end{equation}
This multiplicative formulation ensures that evidence scores contribute only when the answer is already correct—precisely the scenario that Attribution Hallucination concerns. Meanwhile, summing the two evidence scores maintains sufficient within-group reward variance to provide a useful gradient signal.

The group-relative advantages for Eq.~\ref{eq:reward} are computed via within-group normalization. The policy is trained with a clipped importance-weighted objective, using an asymmetric clip range following DAPO \citep{yu2025dapoopensourcellmreinforcement}, along with a KL penalty toward the frozen initial model. Dynamic sampling discards groups whose mean reward falls outside a predefined keep band, as such groups yield near-zero group-relative advantage. The full objective and hyperparameters are detailed in Appendix~\ref{app:impl}, and Algorithm~\ref{alg:grpo} summarizes the training loop. Box recall against manual annotations is used solely for validation and reporting.

\section{Experimental Setup}
\label{sec:setup}

\paragraph{Benchmark.}
We evaluate on the single-document portion of CiteVQA \citep{ma2026citevqabenchmarkingevidenceattribution}. Since some linked PDFs are no longer accessible or differ from the annotated version, we retain only questions where the source PDF is available and the annotated evidence is verifiably present. This filtering is based solely on ground truth and document content, so it favors neither interface, and the dropped documents have text layers at least as rich as the retained ones (Appendix~\ref{app:funnel}); the question-type distribution is also unchanged by the filter (Table~\ref{tab:filtercomp}). The filter retains 719 of 987 original questions (72.9\%), covering 440 PDFs with a median length of 34 pages (max: 182). We call this the verified evaluation set and use it throughout; see Appendix~\ref{app:bench} for composition details.

\paragraph{VLMs.}
We evaluate six instruction-tuned VLMs across four families: Qwen3-VL-8B and Qwen3-VL-30B-A3B \citep{bai2025qwen3vltechnicalreport}, Qwen3.5-9B and Qwen3.5-27B \citep{qwen3.5}, Gemma-3-12B \citep{gemmateam2025gemma3technicalreport}, and Gemma-4-31B \citep{gemmateam2026gemma4technicalreport}. The selection spans dense and MoE architectures, thinking and non-thinking variants, and sizes from 8B to 31B; it also overlaps with CiteVQA's reported models, enabling direct comparison with published coordinate-interface results.

All models process full documents with adaptive per-page resolution capped at one megapixel, matching CiteVQA's protocol (Appendix~\ref{app:impl}). The two thinking models (Qwen3.5-9B and Qwen3.5-27B) and Gemma-4-31B use 4096 output tokens to avoid truncating citations after long outputs, while the other models use 1536; Appendix Table~\ref{tab:models} lists the per-model settings. Greedy decoding is applied throughout. The language interface adds a one-time, cacheable cost per document (the layout parse and about half a minute of GPU time to embed a median of 388 blocks) and a negligible cost per question (Appendix~\ref{app:impl}).

\begin{table}[t]
\centering
\footnotesize
\setlength{\tabcolsep}{3pt}
\begin{tabular}{llp{0.56\columnwidth}}
\toprule
Metric & Source & Definition \\
\midrule
Rec$_{.5}$ & boxes & necessary elements matched at IoU $\ge 0.5$ on the same page, macro-averaged \\
Prec & boxes & annotated elements hit / boxes cited \\
Rel, Ans & judge & evidence relevance and answer quality, 0--5 ($\times 20$) \\
SAA & both & $\mathrm{Ans}\ge4 \wedge (\mathrm{Rel}\ge4 \vee \mathrm{Rec}\ge0.6)$ \\
AH$\downarrow$ & both & $P(\mathrm{SAA}=0 \mid \mathrm{Ans}\ge4)$ \\
fmt & output & responses containing a citation \\
\bottomrule
\end{tabular}
\caption{Summary of the evaluation metrics. Box metrics are judge-free.}
\label{tab:metrics}
\end{table}

\paragraph{Metrics.}
We adopt the CiteVQA evaluation suite (summarized in Table~\ref{tab:metrics}). The annotations distinguish two types of evidence: necessary elements, which every answer must cite, and additional supporting elements. Based on these, we report two core retrieval metrics. Recall (Rec) is the macro-averaged box recall over necessary elements at IoU $\ge 0.5$ on the same page. Precision (Prec) follows the loop semantics of the scorer released with CiteVQA, the number of annotated elements hit divided by the number of boxes cited, and is matched against all annotated evidence rather than the necessary subset, so that citing optional evidence is not penalized (Appendix~\ref{app:residual}). Table~\ref{tab:main} reports these metrics on the subset with parseable judge outputs (Appendix~\ref{app:funnel}); judge-free tables include all valid responses, so recall may differ by up to two points between the two settings. Recall is monotonic in the number of citations, so we report it alongside precision throughout.

We also report judge-based quality metrics. Relevance (Rel) and Answer Quality (Ans) are scores from 0 to 5 provided by a judge (reported scaled by $\times 20$). The Strict Attribution Accuracy (SAA) is defined as
\[
\text{SAA} = \mathbf{1}\big[\mathrm{Ans} \ge 4 \;\wedge\; (\mathrm{Rel} \ge 4 \;\vee\; \mathrm{Rec} \ge 0.6)\big].
\]
From this, we compute the Attribution Hallucination (AH) rate:
\[
\mathrm{AH} = P\!\left(\mathrm{SAA}=0 \;\middle|\; \mathrm{Ans}\ge 4\right),
\]
i.e., the fraction of correct answers that fail to cite appropriate evidence.

We additionally provide multi-threshold recall (same-page, at IoU thresholds from 0.1 to 0.7) and a citation format rate (fmt), defined as the proportion of responses containing any parseable citation. All judge-based evaluations use Gemini-3.5-Flash with the CiteVQA judge prompts (Appendices~\ref{app:prompt-judge-qa} and~\ref{app:prompt-judge-rel}); this judge is a different model from the training reward judge and is not involved in reward computation or retrieval. ANLS as an alternative answer metric is reported in Appendix~\ref{app:anls}.

\paragraph{Training.}
We train Qwen3-VL-8B, chosen as the smallest backbone with reliable citation formatting,
with low-rank adaptation (LoRA) on the language model \citep{hu2021loralowrankadaptationlarge}, keeping the vision tower
frozen, using EasyR1 \citep{zheng2025easyr1}. The training set contains 1{,}584 long-document
questions built from LongDocURL \citep{deng-etal-2025-longdocurl}, rendered as page windows, with
83 held out for validation and no overlap with the evaluation PDFs
(Appendix~\ref{app:impl}). The training reward judge is Qwen3.5-9B, served alongside the trainer. We
sample $n{=}8$ responses per question at temperature 1.0 and train for three epochs (144
steps), evaluating the final checkpoint without selection; the remaining hyperparameters and
the training curve are given in Appendix~\ref{app:impl}.

\begin{table*}[t]
\centering
\small
\begin{tabular}{lcccccc}
\toprule
Model & Rec$_{.5}$ & Prec & Rel & Ans & SAA & AH$\downarrow$ \\
\midrule
\multicolumn{7}{l}{\emph{(a) Coordinate interface: CiteVQA protocol, our reproduction (verified set)}} \\
\midrule
Qwen3-VL-8B       & 0.3 & 0.3 & 15.0 & 59.6 & 1.4 & 97.0 \\
Qwen3-VL-30B-A3B  & 1.3 & 4.6 & 15.7 & 61.9 & 3.8 & 92.1 \\
Qwen3.5-9B        & 4.0 & 5.9 & 21.1 & 89.7 & 11.0 & 87.3 \\
Qwen3.5-27B       & \textbf{8.1} & \textbf{8.6} & 29.1 & \textbf{91.8} & \textbf{16.1} & \textbf{82.1} \\
Gemma-3-12B       & 0.3 & 0.3 & 28.0 & 39.5 & 1.0 & 95.9 \\
Gemma-4-31B       & 7.8 & 6.6 & \textbf{33.0} & 78.7 & 9.2 & 87.1 \\
\midrule
\multicolumn{7}{l}{\emph{(b) Language interface (ours): same backbones as (a), quotes resolved by retrieval (verified set)}} \\
\midrule
Qwen3-VL-8B       & 39.9 & 37.6 & 52.0 & 53.4 & 22.4 & 41.2 \\
Qwen3-VL-30B-A3B  & 34.9 & 35.1 & 42.7 & 60.3 & 24.7 & 47.0 \\
Qwen3.5-9B        & 25.9 & 50.6 & 33.7 & 89.6 & 29.6 & 65.4 \\
Qwen3.5-27B       & 46.9 & \textbf{52.7} & 52.9 & \textbf{90.0} & \textbf{51.5} & 40.5 \\
Gemma-3-12B       & 29.9 & 22.9 & 45.8 & 37.2 & 10.7 & 48.0 \\
Gemma-4-31B       & 44.5 & 45.7 & 56.4 & 77.0 & 42.4 & 38.6 \\
Qwen3-VL-8B + GRPO (ours) & \textbf{51.3} & 26.5 & \textbf{57.4} & 60.4 & 33.8 & \textbf{28.4} \\
\bottomrule
\end{tabular}
\caption{Main results on the verified evaluation set (percentages; Gemini-3.5-Flash judge;
metric definitions in Section~\ref{sec:setup}). Rec$_{.5}$: box recall at IoU 0.5; Prec:
citation precision; Rel, Ans: judge scores ($\times 20$); SAA: strict attributed accuracy;
AH: hallucination among correct answers (lower is better). Bold: best per panel.
Per-condition n is 681 to 719 (Appendix~\ref{app:funnel}).}
\label{tab:main}
\end{table*}

\section{Results}
\label{sec:results}

\subsection{Main comparison}

Table~\ref{tab:main} compares our coordinate-interface reproduction (a) with our language interface (b). Under coordinates, box recall never exceeds 8.1\% and AH ranges from 82.1\% to 97.0\%, even though answer scores reach 91.8. By contrast, the language interface—with the same backbones and inputs—boosts recall to 25.9--46.9\% and reduces AH to 38.6--65.4\%; the GRPO-trained 8B model achieves 51.3\% recall and 33.8\% SAA. Precision follows the same pattern, rising from $\le 8.6\%$ under coordinates to 22.9--52.7\% under language.

Answer scores are stable across interfaces for five of six backbones (within 2.3 points). Qwen3-VL-8B is the exception, dropping 6.2 points under language. The drop concentrates on Quantitative Reasoning questions ($-16.7$), where the worked calculation in the official coordinate prompt leads the model to write out its arithmetic and the JSON format elicits a bare answer \citep{tam-etal-2024-speak}. Gemma-3-12B's low answer scores under both conditions reflect weak Chinese performance (Appendix~\ref{app:breakdown}), not interface effects. All gains in recall, SAA, and AH are significant (paired bootstrap and McNemar, Appendix~\ref{app:stats}).

\begin{figure}[t]
\centering
\includegraphics[width=\columnwidth]{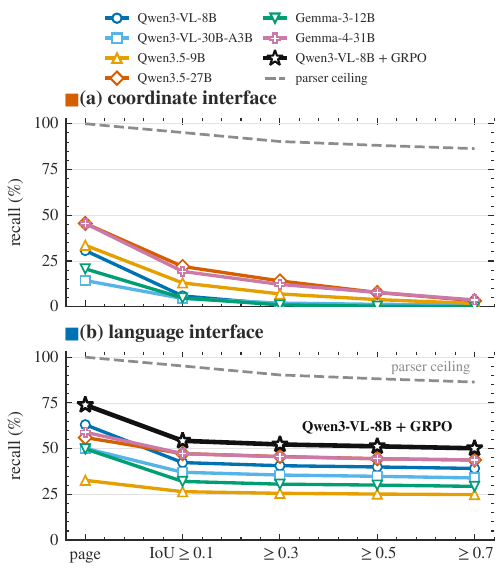}
\caption{Recall of necessary evidence as the overlap requirement tightens, under the
coordinate (a) and language (b) interfaces. Dashed: the
parser ceiling; starred: the GRPO-trained model. Exact values and the
citation-format rate appear in Appendix Table~\ref{tab:threshold}.}
\label{fig:threshold}
\end{figure}

\subsection{Recall by overlap threshold}

Figure~\ref{fig:threshold} shows recall at increasing IoU thresholds. Under coordinates, all backbones exhibit the same pattern: taking Qwen3-VL-8B, recall drops sharply from 30.7\% (page-level) to 5.8\% at IoU 0.1 and 0.0\% at IoU 0.7. Under the language interface, recall is nearly flat across thresholds, ranging for Qwen3-VL-8B from 42.3\% at IoU 0.1 to 39.2\% at IoU 0.7. This flatness is expected: language citations adopt the geometry of parser-proposed blocks, making a correctly matched quote element-precise by design, whereas coordinate boxes miss at element granularity even on the right page.

The dashed line marks the parser ceiling (88.2\% at IoU 0.5, macro-averaged). Both the ceiling and the flat threshold profile replicate with an independent parser (Appendix~\ref{app:crossparser}), confirming robustness.

\begin{table}[t]
\centering
\footnotesize
\setlength{\tabcolsep}{2.0pt}
\begin{tabular}{lcccccc}
\toprule
 & \multicolumn{2}{c}{coordinate} & \multicolumn{3}{c}{retrieval controls} & ours \\
\cmidrule(lr){2-3}\cmidrule(lr){4-6}\cmidrule(lr){7-7}
Backbone & raw & +snap & q & q+a & lex & quotes \\
\midrule
Qwen3-VL-8B      & 0.3 & 9.8  & 27.1 & 38.4 & 28.5 & 39.9 \\
Qwen3-VL-30B     & 1.3 & 5.0  & 27.6 & 37.4 & 24.3 & 34.8 \\
Qwen3.5-9B       & 3.9 & 14.8 & 22.7 & 32.5 & 20.7 & 25.2 \\
Qwen3.5-27B      & 7.8 & 22.9 & 25.8 & 40.5 & 32.4 & 44.5 \\
Gemma-3-12B      & 0.3 & 6.5  & 29.1 & 41.8 & 25.5 & 29.9 \\
Gemma-4-31B      & 7.8 & 23.5 & 27.3 & 42.8 & 33.9 & 44.5 \\
8B + GRPO        & --  & --   & 35.6 & 52.0 & 37.6 & 51.2 \\
\bottomrule
\end{tabular}
\caption{Credit-assignment controls, recall at IoU 0.5 on all 719 valid responses
(Table~\ref{tab:main} uses the judged subset). \textit{raw}: coordinate interface.
\textit{+snap}: each box replaced by the maximum-IoU block on its stated page. \textit{q},
\textit{q+a}: retrieval from the question alone, or the question plus the answer the same
model generated, at that run's citation budget. \textit{lex}: the same quotes resolved
lexically. \textit{quotes}: the full language interface.}
\label{tab:controls}
\end{table}

\begin{figure*}[t]
\centering
\includegraphics[width=\textwidth]{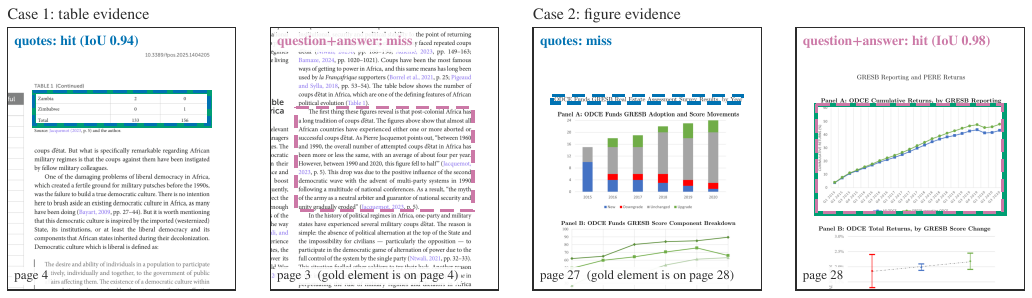}
\caption{Evidence boxes retrieved from the quotes and from the question plus the generated
answer, for the same Qwen3-VL-8B responses, encoder, candidate blocks, and budget. Solid green:
annotated element; dashed: retrieved block. Case 1: the quoted table row resolves to the
annotated table, while question-plus-answer retrieval returns a paragraph on the
previous page. Case 2: the model's quote is a description of a chart and resolves to a heading on the
previous page, while question-plus-answer retrieval returns the chart.}
\label{fig:sidebyside}
\end{figure*}

\subsection{Where the recovery comes from}
\label{sec:controls}

Because the language condition adds three components that the coordinate condition lacks, a parser, a retriever, and an assignment step, Table~\ref{tab:main} shows that the system recovers attribution but not which component is responsible. Table~\ref{tab:controls} presents an ablation study that isolates each component in turn.

\textbf{Granularity alone is not the missing piece.} Snapping each predicted coordinate box to the best-overlapping block on its stated page (``+snap'') gives the coordinate interface the same candidate set and granularity as retrieval. This eliminates the threshold decay seen in coordinates, yet recall remains low, ranging from 5.0\% to 23.5\%, capped by each model's same-page recall.

\textbf{The retriever alone also does not explain the recovery.} We first test retrieval using the question as the query, returning as many blocks per question as the language run originally cited. This budget is set by each model's own citation count, which if anything favors the retrieval control. Under this setting, recall stays below the quote pipeline for every backbone—e.g., by 13 to 19 percentage points for Qwen3-VL-8B, Qwen3.5-27B, and Gemma-4-31B. However, if we append the answer that the same model generated in that run (not the gold answer) to the query, the gap largely closes for those three models and even exceeds the quote pipeline for the other three.

This second comparison is more naturally interpreted as a second reading of the same model output than as an external baseline. The question-only rows measure what the retrieval machinery alone can achieve, while the q+a rows measure the location-relevant information encoded in the model's generated language (Section~\ref{sec:describe}). Importantly, the two readings often retrieve different blocks, and the quotes contribute 11.1 points of coverage that question-plus-answer retrieval never reaches, against 2.4 points for a second query built from the question alone (Appendix~\ref{app:controls}). The strong q+a performance therefore does not render the quote channel redundant. Figure~\ref{fig:sidebyside} shows one example in each direction: a verbatim table row pins the annotated table where the question and answer drift to a topically similar paragraph on another page, and a chart that cannot be quoted is reached through the question and answer.

\textbf{Multimodal encoding contributes beyond string matching.} Lexical and BM25 resolvers operating on the same quotes lose 5 to 14 points compared with the crop-embedding retriever, which also resolves paraphrased quotes rather than relying on literal copying. Full results and quote-fidelity statistics are provided in Appendix~\ref{app:controls}.

\subsection{Effect of region-label-free training}

Training with GRPO using the reward in Eq.~\ref{eq:reward} consistently improves the benchmark's target metrics (Table~\ref{tab:main}, last row). Compared with the Qwen3-VL-8B language baseline, the trained model achieves higher box recall (39.9 → 51.3) and SAA (22.4 → 33.8), while AH drops from 41.2 to 28.4. Same-page recall rises from 63.2 to 73.8, with gains uniform across all IoU thresholds, and the evaluation judge rates the trained model higher on both relevance and answer quality.

These gains come with a trade-off, however. The trained model cites more regions per response (4.6 vs. 2.1 for the baseline), yet citation precision declines from 37.6 to 26.5 and box F1 does not improve (32.2 → 28.8), so part of the recall gain is purchased with additional citations. When the trained model's citations are truncated to the baseline's budget, its recall advantage disappears (Appendix Table~\ref{tab:budget}), so training improves attribution primarily through coverage. We examine the source of these gains and assess whether the extra citations are beneficial or simply noise in Section~\ref{sec:training-fix}.

\section{Discussion}
\label{sec:analysis}

\subsection{An artifact of the coordinate interface}

The main results adjudicate between two hypotheses from the introduction. If models held no location-relevant information, changing the evidence interface should not help; yet retrieval-resolved quotes raise recall by roughly an order of magnitude (Table~\ref{tab:main}), indicating that the coordinate failure is largely an artifact of the output format. The threshold decomposition (Figure~\ref{fig:threshold}) shows what coordinates lose: page identity remains but region geometry does not, as predicted boxes are coarse rectangles around a broadly correct area.

The pattern persists across four families and a four-fold size range; the best coordinate answerer still mis-attributes 82.1\% of its correct answers. Scaling alone does not fix coordinate attribution. The format is nonetheless learnable: reported closed-source results (Appendix Table~\ref{tab:closed}) show Gemini-3.1-Pro reaching 68.9 recall and 76.0 SAA through the same coordinate format, and detection-style coordinate emission succeeds in natural images \citep{chen2022pix2seqlanguagemodelingframework, chen2023shikraunleashingmultimodalllms, you2023ferretrefergroundgranularity}. It is therefore not unlearnable but merely unlearned: open models without dedicated grounding supervision fail with coordinates yet succeed through language, and Section~\ref{sec:controls} traces the recovery to the language they generate.

\subsection{Models already describe locations in language}
\label{sec:describe}

Behavioral evidence comes from the citation-format rate (Appendix Table~\ref{tab:threshold}; see Appendix Figure~\ref{fig:cases} for examples). Under the coordinate prompt, Qwen3-VL-30B-A3B produces a parseable box in only 33.1\% of responses. Manual inspection shows its non-citing outputs are rarely refusals or truncations; rather, the model answers the question and then describes the evidence location verbally—page numbers, section titles, table names. These are metric failures, but they reveal that models asked for coordinates often respond with our proposed language interface instead.

The language interface has its own compliance cost, Qwen3.5-9B emitting parseable citations in only 49.2\% of responses, so format adherence is per-model rather than inherent to either interface.

\subsection{Residual errors and what training fixes}
\label{sec:training-fix}

When the language interface misses an evidence element, the failure can occur at three places: the parser proposes no block that matches the annotation, the model never quotes that evidence, or the correct block is retrieved but lost when the one-to-one assignment gives it to another quote. Appendix~\ref{app:residual} defines these categories and reports their shares before and after training; two readings matter here. First, the parser ceiling is softer than IoU 0.5 suggests, because most unreachable elements are granularity mismatches in which a block does contain the evidence, and we keep IoU 0.5 as primary only for comparability with the published standard. Second, the residual is dominated by evidence the model never quotes rather than by ranking errors, and nearly all of the recall gain from training comes out of exactly that category, which is what the coverage term of the reward is designed to produce.

Are these extra citations noise? The annotations alone cannot say, since precision counts any citation outside a labeled element as wrong. The independent judge, which reads the crops and penalizes irrelevant ones, gives a cleaner signal: under a stricter variant of SAA that drops the recall disjunct and requires $\mathrm{Ans}\ge4 \wedge \mathrm{Rel}\ge4$, the trained model improves from 19.1 to 27.4 (McNemar $p=6\times10^{-7}$, Appendix~\ref{app:stats}). On the 597 questions where the trained model cites more regions than the baseline, judged relevance rises from 52.3 to 60.1 (Appendix~\ref{app:residual}). The extra citations therefore tend to land on supporting material that the annotations do not cover, and the policy improves its quoting rather than exploiting the reward.

\section{Conclusion}

Evidence attribution in visual documents has appeared to be beyond open vision-language
models, a failure read as a capability gap. Expressing evidence as verbatim quotes and resolving them by retrieval recovers
attribution across six open models from four families, and a complementary GRPO-based training improves it further. These findings indicate a practical path to improve attribution without a coordinate interface and without costly region-level supervision in visual document understanding.

\section*{Limitations}

For research scope, our claims are scoped to single-document evidence attribution; multi-document settings add a retrieval dimension we do not study.

For methodology, first, our method does not teach the model to localize on its own: the model only writes quotes, and the parser and the retriever are what turn each quote into a page region. However our pipeline offers ways to get end-to-end region output supervision: its retrieved regions can serve as training labels for such a model, its judge-based reward applies to any model that outputs regions, and a model could simply name one of the parser's blocks instead of writing coordinates \citep{yang2023setofmarkpromptingunleashesextraordinary}. Second, the language interface depends on a layout parser and a cacheable per-document parsing and embedding pass (Appendix~\ref{app:impl}). Its ceiling is therefore the parser's ceiling. Third, for figure evidence, the interface returns the figure's block through its caption, purely graphical evidence with no caption cannot be quoted. Finally, the assignment step always returns a block for every quote, so a hallucinated quote is resolved to a plausible-looking region rather than flagged. Retrieval similarity is informative enough to support abstention (Appendix~\ref{app:abstain}).

For experiment, first, evaluation coverage is bounded by source-PDF availability. The verification filter shrinks the benchmark (Appendix~\ref{app:funnel}), and the retained documents skew toward usable text layers, so performance on heavily scanned collections is untested. Second, all results are single runs with greedy decoding, and the reinforcement learning result is a single seed on one backbone; we report paired uncertainty estimates in Appendix~\ref{app:stats}. Finally, both training and evaluation rely on vision-language judges. The reward judge shapes the training signal, and judged metrics inherit the biases of the judge models. We mitigate this by using the CiteVQA well-designed protocol and an evaluation judge independent of training.

\section*{Acknowledgments}
This work was carried out in AiWo: Human-Centric, AI-Enabled Collaborative Fieldwork
Operations, a Business Finland co-innovation programme under grant agreement No.~69/31/2025.
The computations were performed using computer resources provided by the Aalto University
School of Science ``Science-IT'' project.

We used Claude Code as a coding and writing assistant: implementing training
and evaluation scripts, preparing and analysing data, and assisting with the language of
the manuscript. The research questions, the experimental design, the interpretation of
results, and all claims in this paper are the authors' own; every reported number was
produced by running the released code and verified by the authors.

\bibliography{custom_manual}

\appendix

\section{Implementation Details}
\label{app:impl}

\paragraph{Reproducibility.}
Our repository contains the inference and retrieval pipeline, the evaluation and
control-experiment scripts, and the training configuration. It also provides the question
identifiers of the verified evaluation set with the verification filter, the identifiers of the
training questions, and the trained model, so that every number in the paper can be reproduced
from the public CiteVQA and LongDocURL releases.

\paragraph{Adaptive resolution.}
Every model receives all pages of the document. Each page is rendered at
$\min(10^{6},\, B/n)$ pixels, where $n$ is the page count and $B=1.87\times10^{8}$ is a total
pixel budget chosen so that the rendered document fits the model context. Because the longest
document in the verified set has 182 pages, the per-page resolution is in practice the one-megapixel
cap for every backbone, which matches the standard adaptive scaling of CiteVQA (downscaling to
at most $1024\times1024$ while preserving aspect ratio). Pages are encoded as JPEG images.
Table~\ref{tab:models} lists the checkpoints and the per-model settings.

\begin{table*}[t]
\centering
\small
\begin{tabular}{llccc}
\toprule
Backbone & Hugging Face checkpoint & Context (tokens) & Output budget & Thinking \\
\midrule
Qwen3-VL-8B      & \texttt{Qwen/Qwen3-VL-8B-Instruct}      & 262{,}144 & 1536 & no \\
Qwen3-VL-30B-A3B & \texttt{Qwen/Qwen3-VL-30B-A3B-Instruct} & 262{,}144 & 1536 & no \\
Qwen3.5-9B       & \texttt{Qwen/Qwen3.5-9B}                & 262{,}144 & 4096 & yes \\
Qwen3.5-27B      & \texttt{Qwen/Qwen3.5-27B}               & 262{,}144 & 4096 & yes \\
Gemma-3-12B      & \texttt{google/gemma-3-12b-it}          & 131{,}072 & 1536 & no \\
Gemma-4-31B      & \texttt{google/gemma-4-31B-it}          & 262{,}144 & 4096 & no \\
Qwen3-VL-8B + GRPO & Qwen3-VL-8B with our LoRA adapter     & 262{,}144 & 1536 & no \\
\bottomrule
\end{tabular}
\caption{Per-model inference settings. All models are served with vLLM and share the prompts,
the one-megapixel page rendering, and temperature 0 under both interfaces.}
\label{tab:models}
\end{table*}

\paragraph{Retrieval implementation.}
Semantic blocks are taken from the MinerU parse of each PDF (version 3.2.3, pipeline backend, with table and formula recognition disabled; median 388 blocks per document).
Block crops are rendered at 150 dpi and embedded once per document with Qwen3-VL-Embedding-2B;
the embeddings are cached on disk. Quotes are embedded as text with the same encoder.
Similarities are cosine over unit-normalized embeddings, and the one-to-one assignment is
solved with the Hungarian algorithm (SciPy \texttt{linear\_sum\_assignment}); when a response contains more quotes than the document has
blocks, each quote falls back to its highest-similarity block.

\paragraph{Training data construction.}
LongDocURL \citep{deng-etal-2025-longdocurl} provides long-document QA pairs with gold answers and
evidence-page annotations. For each question we render a window of up to 30 pages around the
annotated evidence pages at a budget of ten megapixels per window with the same one-megapixel cap per page as evaluation (a 30-page window is thus rendered at about a third of a megapixel per page), which keeps long
documents tractable for rollout sampling; the evidence-page annotation is used only to place
the window, never in the reward. We first sample eight responses per question from the
base model on 2{,}252 candidate questions and discard the questions on which it already
obtains a consistently high reward, since they carry no learning signal. The resulting set contains 1{,}584 training questions over 349
documents, with 83 questions held out for validation. A whole-text document fingerprint check
confirms that no training document overlaps the CiteVQA evaluation PDFs. The rollout prompt is
the same natural-language citation contract as evaluation (Appendix~\ref{app:prompt-nl}).

\paragraph{Optimization objective.}
For a question $q$ over document $D$, the rollout policy $\pi_{\theta_{\mathrm{old}}}$ samples
a group of $n$ responses $\{o_i\}_{i=1}^{n}$, and the scalar rewards $r_i$ of
Eq.~\ref{eq:reward} are normalized within the group,
\begin{equation}
\label{eq:adv}
\hat{A}_i \;=\; \frac{r_i - \operatorname{mean}\big(\{r_j\}_{j=1}^{n}\big)}
{\operatorname{std}\big(\{r_j\}_{j=1}^{n}\big) + \epsilon},
\end{equation}
so that every token of $o_i$ shares the advantage $\hat{A}_i$. With the token-level importance
ratio $\rho_{i,t}(\theta)=\pi_\theta(o_{i,t}\mid q, D, o_{i,<t})\,/\,
\pi_{\theta_{\mathrm{old}}}(o_{i,t}\mid q, D, o_{i,<t})$, training maximizes
\begin{equation}
\label{eq:grpo}
\begin{split}
\mathcal{J}(\theta) = \mathbb{E}\bigg[\frac{1}{n}\sum_{i=1}^{n}\frac{1}{|o_i|}\sum_{t}
\min\Big(\rho_{i,t}\hat{A}_i,\; \qquad\\
\operatorname{clip}\big(\rho_{i,t},\, 1-\varepsilon_{\mathrm{l}},\,
1+\varepsilon_{\mathrm{h}}\big)\hat{A}_i\Big)\bigg]
- \beta\, \mathbb{D}_{\mathrm{KL}}\big[\pi_\theta \,\|\, \pi_{\mathrm{ref}}\big],
\end{split}
\end{equation}
where $\pi_{\mathrm{ref}}$ is the frozen initial model, the KL divergence is computed with the
low-variance estimator of \citet{shao2024deepseekmathpushinglimitsmathematical}, and the clip range is asymmetric
with $\varepsilon_{\mathrm{h}}>\varepsilon_{\mathrm{l}}$ following DAPO \citep{yu2025dapoopensourcellmreinforcement}.

\begin{algorithm}[t]
\caption{Region-label-free GRPO for evidence attribution}
\label{alg:grpo}
\begin{algorithmic}[1]
\For{each training question $(q, D, y^{*})$}
  \State sample $n$ responses $o_1,\dots,o_n \sim \pi_{\theta_{\mathrm{old}}}(\cdot \mid q, D)$
  \For{each response $o_i$}
    \State parse $o_i$ into answer $y_i$ and quotes $U_i$
    \State $C_i \gets \mathcal{A}\big(f(U_i), f(\mathcal{B}(D))\big)$ \Comment{retrieval, Sec.~\ref{sec:retrieval}}
    \State judge $(q, y^{*}, y_i,$ crops of $C_i)$ $\to a_i, e_{\mathrm{rel},i}, e_{\mathrm{cov},i}$
    \State $r_i \gets (a_i/5)\cdot(e_{\mathrm{rel},i}+e_{\mathrm{cov},i})/10$ \Comment{Eq.~\ref{eq:reward}}
  \EndFor
  \If{$\operatorname{mean}_i r_i \notin [0.01, 0.99]$} discard group \Comment{dynamic sampling}
  \EndIf
  \State $\hat{A}_i \gets (r_i - \operatorname{mean} r)/(\operatorname{std} r + \epsilon)$ \Comment{Eq.~\ref{eq:adv}}
  \State update $\theta$ by maximizing Eq.~\ref{eq:grpo}
\EndFor
\end{algorithmic}
\end{algorithm}

\begin{figure}[htbp]
\centering
\includegraphics[width=\columnwidth]{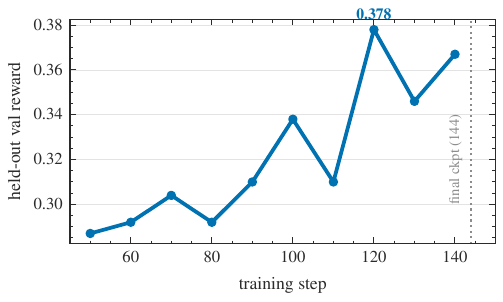}
\caption{Held-out validation reward during GRPO training. The final checkpoint (step 144) is
evaluated without selection.}
\label{fig:valcurve}
\end{figure}

\begin{figure}[htbp]
\centering
\includegraphics[width=\columnwidth]{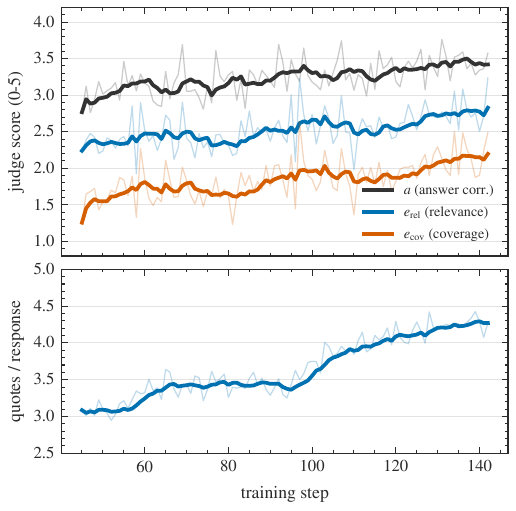}
\caption{Training dynamics of the reward components (top; judge scores on the 0--5 scale) and
of the citing behavior (bottom; quotes per response), shown from step 45 onward; per-step
metrics for steps 1 to 44 are unavailable because a cluster requeue overwrote the metrics file
before it was archived. Bold lines are 7-step moving averages over the per-step values (faint).
The coverage component and the number of quotes rise the most, consistent with the training
gain being driven by citing previously omitted evidence.}
\label{fig:rewarddyn}
\end{figure}

\paragraph{Training hyperparameters.}
LoRA rank 64, alpha 64, applied to all linear layers of the language model, with the vision
tower frozen. Rollouts use $n{=}8$ samples per question at temperature 1.0, maximum prompt
length 48{,}000 tokens, and maximum response length 768 tokens. The keep band for group
filtering is $[0.01, 0.99]$ on the group mean reward. The learning rate is $10^{-5}$ with a
constant schedule, rollout batch 32, global batch 16, three epochs, 144 optimizer steps. The
training reward judge is Qwen3.5-9B served with vLLM alongside the trainer. The objective of
Eq.~\ref{eq:grpo} uses clip range $(\varepsilon_{\mathrm{l}}, \varepsilon_{\mathrm{h}}) =
(0.2, 0.28)$, KL coefficient $\beta = 0.01$ against the frozen initial model with the
low-variance estimator, and group-normalization constant $\epsilon = 10^{-6}$ in
Eq.~\ref{eq:adv}. Figure~\ref{fig:valcurve} shows the held-out validation reward, and
Figure~\ref{fig:rewarddyn} tracks the three reward components together with the number of
quotes per response. Coverage starts lowest and rises the most (1.4 to 2.5), relevance gains
about one point, answer correctness moves least, and the model cites roughly one additional
quote per response by the end of training; these dynamics match the element-level
decomposition of Figure~\ref{fig:residual}, where most of the recall gain comes from evidence
that was previously not cited.

\paragraph{Training and evaluation cost.}
Reinforcement learning ran as four chained cluster jobs totaling 28.4 wall-clock hours on three
H200 GPUs each (two for the trainer and one serving the reward judge and the embedder), or
approximately 85 GPU-hours. Evaluating the 13-condition matrix (six backbones under two
interfaces, plus the trained model) took approximately 60 single-H200 GPU-hours of inference,
dominated by the two thinking models with their 4096-token output budgets. Diagnostic analyses
(retrieval headroom and parser ceiling) add under two GPU-hours. At inference, the language
interface adds a one-time per-document cost, the layout parse and one embedding per block
(median 388 blocks, on the order of seconds for parsing and half a minute of GPU time for
embeddings per document), which is cacheable across questions of the same document, and a
per-question cost of embedding the quotes and solving one assignment, which is negligible;
the coordinate interface adds no overhead. Judging costs are listed below.

\paragraph{Evaluation judging.}
The evaluation judge is Gemini-3.5-Flash, called through the batch API with thinking disabled,
temperature 0, and a 1024-token output limit. For each response, the answer-accuracy judge
receives the question, the gold answer, and the model answer, and the relevance judge
additionally receives the crops of the cited regions rendered at 150 dpi (at most six per
response). Judging the full 13-condition matrix took roughly 18{,}950 requests, 40.9M input tokens and 1.6M output tokens in total.

\paragraph{Benchmark construction.}
\label{app:bench}
Starting from the CiteVQA single-document questions, we keep a question when (i) its source PDF
resolves and downloads, (ii) all annotated evidence pages exist in the downloaded file, and
(iii) the annotated textual evidence is verifiably present at the annotated location,
established by extracting the text under each ground-truth box and matching it against the
annotated content. The checks inspect only ground-truth annotations and never model outputs.
Table~\ref{tab:bench} summarizes the resulting set. Question types follow the CiteVQA taxonomy:
Complex Synthesis questions combine several elements, often across pages; Multimodal Parsing
questions target a specific table or figure; Factual Retrieval questions look up a single fact;
Quantitative Reasoning questions require numeric computation over document content. The
non-textual share is the fraction of necessary ground-truth elements whose annotated type is a
table or a figure.

\begin{table}[t]
\centering
\small
\begin{tabular}{lr}
\toprule
Property & Value \\
\midrule
Questions & 719 \\
Source PDFs & 440 \\
Language (en / zh) & 386 / 333 \\
Complex Synthesis & 394 (55\%) \\
Multimodal Parsing & 136 \\
Factual Retrieval & 122 \\
Quantitative Reasoning & 67 \\
Pages (median / p90 / max) & 34 / 104 / 182 \\
Non-textual evidence & $\sim$30\% \\
\bottomrule
\end{tabular}
\caption{Composition of the verified single-document evaluation set.}
\label{tab:bench}
\end{table}

\begin{table*}[t]
\centering
\small
\begin{tabular}{llcccccc}
\toprule
Backbone & Interface & page & @0.1 & @0.3 & @0.5 & @0.7 & fmt \\
\midrule
\multicolumn{2}{l}{\emph{Parser ceiling (upper bound for retrieval)}} & 100.0 & 95.2 & 90.3 & 88.2 & 86.4 & -- \\
\midrule
\multirow{2}{*}{Qwen3-VL-8B} & coord. & 30.7 & 5.8 & 1.2 & 0.3 & 0.0 & 61.5 \\
 & lang. & 63.2 & 42.3 & 40.7 & 39.9 & 39.2 & 97.5 \\
\multirow{2}{*}{Gemma-3-12B} & coord. & 20.6 & 4.8 & 1.0 & 0.3 & 0.0 & 95.1 \\
 & lang. & 49.8 & 32.0 & 30.5 & 29.9 & 29.3 & 99.7 \\
\multirow{2}{*}{Qwen3.5-9B} & coord. & 33.6 & 13.0 & 7.0 & 3.9 & 1.8 & 63.3 \\
 & lang. & 32.7 & 26.5 & 25.6 & 25.2 & 24.9 & 49.2 \\
\multirow{2}{*}{Qwen3-VL-30B-A3B} & coord. & 14.3 & 4.5 & 1.9 & 1.3 & 1.1 & 33.1 \\
 & lang. & 50.2 & 37.1 & 35.6 & 34.8 & 34.0 & 75.8 \\
\multirow{2}{*}{Gemma-4-31B} & coord. & 45.5 & 19.3 & 12.2 & 7.8 & 3.7 & 99.4 \\
 & lang. & 59.0 & 47.3 & 45.6 & 44.5 & 43.8 & 88.3 \\
\multirow{2}{*}{Qwen3.5-27B} & coord. & 45.5 & 22.0 & 14.1 & 7.8 & 3.3 & 86.6 \\
 & lang. & 56.1 & 47.3 & 45.6 & 44.5 & 43.8 & 76.1 \\
\midrule
Qwen3-VL-8B + GRPO & lang. & \textbf{73.8} & \textbf{54.2} & \textbf{52.2} & \textbf{51.2} & \textbf{50.1} & 97.9 \\
\bottomrule
\end{tabular}
\caption{Multi-threshold recall of necessary evidence (same-page recall, then same page with
IoU $\ge t$) and citation-format rate, over all valid responses (n=719). Qwen3.5-27B and
Gemma-4-31B coincide to one decimal at every IoU threshold; they differ at the second decimal. The parser ceiling is the recall achievable if retrieval
were perfect, i.e., matching against every semantic block of the document.}
\label{tab:threshold}
\end{table*}

\paragraph{Protocol checks.}
Three properties support the fairness of the comparison. The coordinate condition uses the
CiteVQA prompt verbatim (including its minor typographical errors, which we preserve
unchanged), and our coordinate numbers fall in the same regime as the published
single-document scores for shared models. The two thinking models and Gemma-4-31B receive a 4096-token output
budget because long outputs truncate final citations at smaller budgets; this
correction raises coordinate scores. The verification filter used to build the verified evaluation set inspects
only ground-truth annotations, never model outputs.

\paragraph{Closed-source reference points.}
Table~\ref{tab:closed} reproduces the closed-source rows of the CiteVQA main table, which we
do not rerun. They serve one purpose in our argument, discussed in
Section~\ref{sec:analysis}: systems trained at scale do reach useful attribution accuracy
through the coordinate interface, so the failure we diagnose in open models is a matter of
missing supervision rather than of an unlearnable output format. The numbers are not
comparable to Table~\ref{tab:main} in absolute terms, since they come from the full
single-document split scored by a different judge, and CiteVQA does not report AH.

\begin{table}[t]
\centering
\small
\setlength{\tabcolsep}{4pt}
\begin{tabular}{lcccc}
\toprule
System & Rec$_{.5}$ & Rel & Ans & SAA \\
\midrule
Gemini-3.1-Pro-Preview  & 68.9 & 82.6 & 86.7 & 76.0 \\
Gemini-3-Flash-Preview  & 49.5 & 76.8 & 85.3 & 69.3 \\
Gemini-2.5-Pro          & 31.5 & 61.7 & 83.0 & 49.4 \\
GPT-5.4                 & 35.9 & 69.8 & 87.6 & 61.7 \\
GPT-5.2                 & 20.9 & 54.9 & 71.4 & 32.6 \\
Qwen3.6-Plus            & 9.8  & 26.7 & 87.1 & 20.2 \\
Seed2.0-Pro             & 35.8 & 60.8 & 82.9 & 51.9 \\
GLM-5V-Turbo            & 18.3 & 31.2 & 50.0 & 14.1 \\
\bottomrule
\end{tabular}
\caption{Closed-source coordinate-interface results reproduced from the CiteVQA main table
\citep{ma2026citevqabenchmarkingevidenceattribution}: their full single-document split, Qwen3-VL-235B judge, temperature
1.0. These numbers are reference points rather than direct comparisons with
Table~\ref{tab:main}, which uses the verified evaluation set and a different judge; CiteVQA
does not report AH.}
\label{tab:closed}
\end{table}

\section{Benchmark Filtering Funnel}
\label{app:funnel}
The CiteVQA validation release contains 987 single-document questions. The verification filter
removes 117 questions whose source PDF does not resolve or is not a valid PDF, 4 whose
annotated evidence pages lie outside the downloaded file, 131 whose annotated evidence text
cannot be found at the annotated location, and 16 whose document differs from the annotated
version and cannot be byte-verified, leaving 719 questions (72.9 percent). To test whether the
text-verification criterion skews the benchmark toward born-digital documents, we measured the
share of pages with a substantive extractable text layer (more than 200 characters) for both
groups. Retained documents average a 0.78 text-layer share and include 45 scan-like documents
(10.2 percent) with text on fewer than a quarter of their pages; documents dropped by the
text-mismatch criterion average a 0.84 share with 3.8 percent scan-like. The filter therefore
does not select for parser-friendly documents; the text-mismatch drops reflect
annotation-document disagreements rather than scanned-document exclusion. Per-condition n in
Table~\ref{tab:main} varies between 681 and 719 because judge responses that contain no
parseable score tag (13 of the 9{,}347 answer calls and 240 of the 9{,}347 relevance calls,
13 conditions over 719 questions) exclude their question from that condition.

Table~\ref{tab:filtercomp} compares the composition of the benchmark before and after
filtering: the question-type distribution, the number of necessary elements per question, and
the share of non-textual evidence stay close to the original. Table~\ref{tab:scan}
slices the verified set by the text-layer share of the document. The language interface reaches
similar recall on the 82 questions over scan-like documents and on born-digital ones.

\begin{table}[t]
\centering
\footnotesize
\setlength{\tabcolsep}{3pt}
\begin{tabular}{lccc}
\toprule
 & original & verified & dropped \\
\midrule
Questions & 987 & 719 & 268 \\
Complex Synthesis & 55\% & 55\% & 55\% \\
Multimodal Parsing & 18\% & 19\% & 16\% \\
Factual Retrieval & 19\% & 17\% & 23\% \\
Quantitative Reasoning & 9\% & 9\% & 6\% \\
Chinese & 50\% & 46\% & 61\% \\
Necessary elements / question & 1.42 & 1.44 & 1.39 \\
Non-textual necessary evidence & 31\% & 33\% & 26\% \\
\bottomrule
\end{tabular}
\caption{Composition of the CiteVQA single-document questions before filtering, of the
verified evaluation set, and of the dropped questions.}
\label{tab:filtercomp}
\end{table}

\begin{table}[t]
\centering
\footnotesize
\setlength{\tabcolsep}{3pt}
\begin{tabular}{lccc}
\toprule
Condition & scan-like & partial & born-digital \\
 & (82) & (96) & (541) \\
\midrule
Qwen3-VL-8B coord.  & 0.6 & 0.0 & 0.2 \\
Qwen3-VL-30B coord. & 1.2 & 1.0 & 1.3 \\
Qwen3.5-9B coord.   & 3.4 & 2.6 & 4.2 \\
Qwen3.5-27B coord.  & 6.7 & 7.8 & 8.0 \\
Gemma-3-12B coord.  & 1.2 & 0.0 & 0.2 \\
Gemma-4-31B coord.  & 9.5 & 8.7 & 7.4 \\
\midrule
Qwen3-VL-8B lang.   & 44.0 & 30.4 & 41.0 \\
Qwen3-VL-30B lang.  & 31.7 & 36.2 & 35.1 \\
Qwen3.5-9B lang.    & 25.2 & 26.9 & 24.9 \\
Qwen3.5-27B lang.   & 41.5 & 42.0 & 45.4 \\
Gemma-3-12B lang.   & 35.5 & 33.6 & 28.5 \\
Gemma-4-31B lang.   & 44.0 & 40.9 & 45.2 \\
8B + GRPO lang.     & 52.6 & 44.9 & 52.1 \\
\bottomrule
\end{tabular}
\caption{Recall at IoU 0.5 by document text layer (number of questions in parentheses).
Scan-like: fewer than a quarter of the pages carry a substantive text layer; born-digital: at
least three quarters.}
\label{tab:scan}
\end{table}

\section{Per-Language and Per-Type Results}
\label{app:breakdown}
The interface effect and the training gain hold in both benchmark languages
(Table~\ref{tab:perlang}); Chinese questions score at least as well as English ones under the
language interface. At the level of individual evidence elements, the language interface
recovers tables nearly as well as running text, while figures remain the hardest category,
consistent with their caption-based anchoring; coordinate rows stay near zero for most
backbones in every category (Table~\ref{tab:pertype}). By question type
(Table~\ref{tab:qtype}), the language interface is above the coordinate interface for every
backbone and every type. Complex Synthesis, the type whose evidence is spread over several
elements, ranks first or second for every language run; Quantitative Reasoning is high for most
backbones but is the lowest type for Qwen3.5-9B and Gemma-4-31B.

\begin{table}[t]
\centering
\footnotesize
\setlength{\tabcolsep}{3.5pt}
\begin{tabular}{lcccc}
\toprule
Condition & CS & MP & FR & QR \\
\midrule
Qwen3-VL-8B coord.  & 0.4 & 0.2 & 0.0 & 0.0 \\
Qwen3-VL-30B coord. & 0.5 & 2.2 & 0.8 & 4.5 \\
Qwen3.5-9B coord.   & 4.1 & 2.2 & 1.9 & 9.5 \\
Qwen3.5-27B coord.  & 9.0 & 5.4 & 4.1 & 12.4 \\
Gemma-3-12B coord.  & 0.3 & 0.7 & 0.0 & 0.0 \\
Gemma-4-31B coord.  & 8.8 & 8.2 & 6.6 & 3.7 \\
\midrule
Qwen3-VL-8B lang.   & 43.5 & 34.8 & 31.5 & 44.8 \\
Qwen3-VL-30B lang.  & 38.5 & 28.5 & 28.3 & 38.1 \\
Qwen3.5-9B lang.    & 25.9 & 27.9 & 24.5 & 17.2 \\
Qwen3.5-27B lang.   & 47.7 & 38.7 & 39.5 & 46.3 \\
Gemma-3-12B lang.   & 36.0 & 23.8 & 20.2 & 24.6 \\
Gemma-4-31B lang.   & 47.9 & 42.6 & 39.3 & 37.3 \\
8B + GRPO lang.     & 55.5 & 43.8 & 44.4 & 53.2 \\
\bottomrule
\end{tabular}
\caption{Recall at IoU 0.5 by question type (n=719). CS: Complex Synthesis (394 questions);
MP: Multimodal Parsing (136); FR: Factual Retrieval (122); QR: Quantitative Reasoning (67).}
\label{tab:qtype}
\end{table}

\begin{table*}[t]
\centering
\small
\begin{tabular}{lcccccc}
\toprule
 & \multicolumn{3}{c}{English (386)} & \multicolumn{3}{c}{Chinese (333)} \\
\cmidrule(lr){2-4}\cmidrule(lr){5-7}
Condition & Rec$_{.5}$ & SAA & AH & Rec$_{.5}$ & SAA & AH \\
\midrule
Qwen3-VL-8B coord. & 0.4 & 1.0 & 97.9 & 0.1 & 1.8 & 95.9 \\
Qwen3.5-27B coord. & 6.0 & 11.7 & 86.8 & 10.4 & 21.0 & 76.9 \\
Gemma-4-31B coord. & 7.7 & 10.6 & 85.5 & 7.9 & 7.5 & 89.0 \\
Qwen3-VL-8B lang.  & 37.2 & 20.7 & 41.6 & 43.2 & 24.3 & 40.9 \\
Qwen3.5-27B lang.  & 39.9 & 44.8 & 47.1 & 55.1 & 59.4 & 33.1 \\
Gemma-4-31B lang.  & 44.2 & 46.1 & 35.3 & 44.7 & 38.1 & 42.8 \\
8B + GRPO lang.    & 50.5 & 34.2 & 27.6 & 52.2 & 33.3 & 29.3 \\
\bottomrule
\end{tabular}
\caption{Per-language results for representative conditions (the remaining backbones follow the
same pattern). Judged subsets per condition.}
\label{tab:perlang}
\end{table*}

\begin{table}[t]
\centering
\small
\setlength{\tabcolsep}{4pt}
\begin{tabular}{lccc}
\toprule
Condition & text & table & figure \\
\midrule
Qwen3-VL-8B coord. & 0.3 & 0.4 & 0.0 \\
Qwen3.5-27B coord. & 5.7 & 10.0 & 11.3 \\
Gemma-4-31B coord. & 7.9 & 6.3 & 12.7 \\
Qwen3-VL-8B lang.  & 38.1 & 35.4 & 9.9 \\
Qwen3.5-27B lang.  & 43.8 & 35.1 & 12.7 \\
Gemma-4-31B lang.  & 44.4 & 34.7 & 19.7 \\
8B + GRPO lang.    & 51.1 & 44.6 & 19.7 \\
\bottomrule
\end{tabular}
\caption{Element-level recall at IoU 0.5 by evidence type (685 text, 271 table, 71 figure
elements; text includes titles and lists, table includes captions and footnotes, figure
includes image captions).}
\label{tab:pertype}
\end{table}

\section{Statistical Tests}
\label{app:stats}
For every backbone we pair the two interface conditions on the questions judged in both, and
for GRPO we pair against its base model. Differences in recall are tested with a paired
bootstrap over questions (10{,}000 resamples, fixed seed), SAA with an exact McNemar test on
discordant pairs, and AH with a paired bootstrap of the conditional rate; we also recompute AH
on the intersection of questions answered correctly (Ans $\ge 4$) under both conditions, which
removes any composition effect from differing denominators. Table~\ref{tab:stats} shows that
every contrast is significant and that intersection-AH is within two points of the
unconditional AH throughout.

\begin{table*}[t]
\centering
\small
\begin{tabular}{lcccccc}
\toprule
Pair (NL vs. coord.) & n & $\Delta$Rec$_{.5}$ [95\% CI] & $\Delta$SAA [95\% CI] & McNemar $p$ &
AH pair & inters.-AH \\
\midrule
Qwen3-VL-8B  & 717 & $+39.8$ [36.4, 43.2] & $+21.1$ [18.0, 24.3] & $6.0\times10^{-40}$ & 41.0 / 97.0 & 38.5 / 96.3 \\
Qwen3-VL-30B & 717 & $+33.7$ [30.5, 37.0] & $+20.8$ [17.6, 24.1] & $6.7\times10^{-34}$ & 47.1 / 92.1 & 43.0 / 91.6 \\
Qwen3.5-9B   & 679 & $+21.6$ [18.2, 25.0] & $+18.1$ [14.1, 22.1] & $4.6\times10^{-18}$ & 66.0 / 87.4 & 65.8 / 87.1 \\
Qwen3.5-27B  & 656 & $+39.4$ [35.3, 43.4] & $+36.4$ [31.7, 41.2] & $3.2\times10^{-44}$ & 40.4 / 82.6 & 40.0 / 82.5 \\
Gemma-3-12B  & 719 & $+29.7$ [26.5, 32.8] & $+9.7$ [7.5, 12.1]  & $1.0\times10^{-17}$ & 48.0 / 95.9 & 45.8 / 96.3 \\
Gemma-4-31B  & 718 & $+36.7$ [32.9, 40.4] & $+33.3$ [29.2, 37.2] & $3.4\times10^{-49}$ & 38.6 / 87.1 & 37.1 / 87.2 \\
\midrule
GRPO vs. base & 716 & $+11.2$ [8.1, 14.3] & $+11.3$ [8.2, 14.5] & $4.0\times10^{-12}$ & 28.4 / 41.0 & 27.8 / 39.2 \\
\bottomrule
\end{tabular}
\caption{Paired significance tests (NL: language interface). CI: paired bootstrap over
questions (10k resamples). AH pair: language / coordinate (for the last row, GRPO / base). Intersection-AH restricts both
conditions to questions with Ans $\ge 4$ under both.}
\label{tab:stats}
\end{table*}

Under the stricter SAA variant of Section~\ref{sec:training-fix} that requires judged
relevance alone ($\mathrm{Ans}\!\ge\!4 \wedge \mathrm{Rel}\!\ge\!4$, no recall
disjunct), the GRPO model improves over its base from 19.1 to 27.4 (discordant pairs $b=99$,
$c=40$, McNemar $p=6.0\times10^{-7}$), and the corresponding hallucination rate falls from
50.0 to 42.0.

\section{Cross-Parser Check}
\label{app:crossparser}
The layout parser and the CiteVQA annotations could in principle share segmentation
conventions, which would inflate the parser ceiling and the flat threshold profile of the
language interface. To test this, we re-parsed all 440 documents with Docling
\citep{auer2024doclingtechnicalreport}, an independently developed layout parser that shares no lineage with the
annotation tooling, and repeated the judge-free geometric evaluation. The Docling ceiling is
100.0 (page), 95.9 (IoU 0.1), 91.0 (0.3), 87.8 (0.5), and 79.8 (0.7), against 100.0, 95.2,
90.3, 88.2, and 86.4 for the parser used in the main experiments, although Docling segments
about twice as finely (median 884 blocks per document against 388). The near-identical
ceiling at the standard threshold indicates that the reachability of the annotated elements
is a property of the documents rather than of one parser's segmentation conventions.
Table~\ref{tab:crossparser} repeats the language-interface evaluation with the model quotes
held fixed and only the parser swapped. Recall is 3 to 6 points below the main results,
consistent with the finer segmentation, but the profile that carries the paper's argument is
unchanged: recall stays nearly flat from IoU 0.1 to 0.7, the ordering of the backbones is
preserved, and the training gain persists (44.9 against 35.9 for the base model at IoU 0.5).
The flat threshold profile is therefore not an artifact of one parser's segmentation
conventions.

\begin{table}[t]
\centering
\footnotesize
\setlength{\tabcolsep}{2.8pt}
\begin{tabular}{lccccc}
\toprule
Backbone & page & @0.1 & @0.3 & @0.5 & @0.7 \\
\midrule
Qwen3-VL-8B      & 63.6 & 38.6 & 37.0 & 35.9 & 32.7 \\
Qwen3-VL-30B     & 53.1 & 32.3 & 31.2 & 29.6 & 26.6 \\
Qwen3.5-9B       & 33.8 & 23.5 & 21.9 & 20.9 & 18.6 \\
Qwen3.5-27B      & 61.2 & 43.6 & 41.1 & 39.9 & 35.5 \\
Gemma-3-12B      & 53.7 & 27.6 & 25.8 & 24.7 & 22.2 \\
Gemma-4-31B      & 63.2 & 42.2 & 39.7 & 38.4 & 34.7 \\
8B + GRPO        & 75.2 & 48.8 & 46.4 & 44.9 & 40.6 \\
\bottomrule
\end{tabular}
\caption{Language-interface recall with Docling blocks in place of the main parser, on
identical model quotes. The flat threshold profile and the training gain replicate under the
independent parser.}
\label{tab:crossparser}
\end{table}

\section{Credit-Assignment Controls}
\label{app:controls}

\paragraph{Coordinate-to-block snapping.}
Table~\ref{tab:snap} reports the full multi-threshold results of the snapping control from
Section~\ref{sec:controls}. Snapping removes the threshold decay, which confirms that the raw
coordinate collapse between IoU 0.1 and 0.7 is a granularity phenomenon, but the snapped recall
remains bounded by each model's same-page recall and far below the language interface.

\begin{table}[t]
\centering
\footnotesize
\setlength{\tabcolsep}{2.8pt}
\begin{tabular}{lcccccc}
\toprule
Backbone & var. & page & @0.1 & @0.3 & @0.5 & @0.7 \\
\midrule
\multirow{2}{*}{Qwen3-VL-8B} & raw & 30.7 & 5.8 & 1.2 & 0.3 & 0.0 \\
 & snap & 30.7 & 10.6 & 9.9 & 9.8 & 9.4 \\
\multirow{2}{*}{Qwen3-VL-30B} & raw & 14.3 & 4.5 & 1.9 & 1.3 & 1.1 \\
 & snap & 14.3 & 5.6 & 5.1 & 5.0 & 4.9 \\
\multirow{2}{*}{Qwen3.5-9B} & raw & 33.6 & 13.0 & 7.0 & 3.9 & 1.8 \\
 & snap & 33.6 & 16.6 & 15.1 & 14.8 & 14.2 \\
\multirow{2}{*}{Qwen3.5-27B} & raw & 45.5 & 22.0 & 14.1 & 7.8 & 3.3 \\
 & snap & 45.5 & 25.1 & 23.6 & 22.9 & 22.2 \\
\multirow{2}{*}{Gemma-3-12B} & raw & 20.6 & 4.8 & 1.0 & 0.3 & 0.0 \\
 & snap & 20.6 & 7.2 & 6.5 & 6.5 & 6.4 \\
\multirow{2}{*}{Gemma-4-31B} & raw & 45.5 & 19.3 & 12.2 & 7.8 & 3.7 \\
 & snap & 45.5 & 25.8 & 24.3 & 23.5 & 22.9 \\
\bottomrule
\end{tabular}
\caption{Snapping control, full thresholds (n=719). Each predicted coordinate box is snapped to
the maximally overlapping semantic block on its stated page, with a nearest-block fallback when
no block overlaps and no change when the page has no parsed blocks.}
\label{tab:snap}
\end{table}

\paragraph{Retrieval-only baselines.}
With the question alone as the query and a fixed budget of $k$ blocks per question,
recall at IoU 0.5 is 18.9 at $k{=}1$, 33.1 at $k{=}3$, and 39.9 at $k{=}5$; the corresponding
same-page recalls are 34.9, 54.1, and 65.9. These fixed-budget numbers are not comparable to
the language runs, which average between 1.1 and 3.3 citations per response; the budget-matched
comparison appears in Table~\ref{tab:controls}. Splitting the budget-matched comparison by
answer correctness sharpens the reading. On correctly answered questions (Ans $\ge 4$), quote
recall exceeds the question-plus-answer control by 4.5 to 7.1 points (for Qwen3-VL-8B, 52.1
against 45.0, over a question-only floor of 28.3), while on incorrectly answered questions
both controls fall toward the floor and the quote margin vanishes. Quotes and answers
therefore rise and fall together as expressions of the same underlying evidence
identification, and on the same correctly answered questions the coordinate interface stays
below 8.1 recall, which is the dissociation that the main results quantify as AH.

\paragraph{Complementarity of quotes and answers.}
The similar aggregate recall of the quotes and q+a rows in Table~\ref{tab:controls} hides
that the two readings succeed on different questions: their per-question recalls correlate at
only 0.39 to 0.54 (Pearson) across backbones and differ on 30 percent of question and
backbone pairs. Table~\ref{tab:union} scores each question by
the better of the two readings, an oracle union that diagnoses how disjoint the two signals
are rather than describing a deployable system. Pooled over the seven language runs, the
union reaches 51.9 recall against 40.8 for q+a alone. As a control for generic query
diversity, the union of question-only retrieval with q+a adds only 2.4 points over q+a,
while adding the quotes adds 11.1, so the extra coverage comes from what the quotes say and
not from issuing a second query. For questions with several necessary elements the
per-question maximum is a lower bound on the true union; on the 505 questions with a single
necessary element, where the union is exact, the pattern is unchanged (for Gemma-4-31B, 61.2
against 50.5 for the diversity control). Both readings use the same encoder, candidate set,
and per-question citation budget throughout.

\begin{table}[t]
\centering
\footnotesize
\setlength{\tabcolsep}{3.2pt}
\begin{tabular}{lcccc}
\toprule
Backbone & quotes & q+a & quotes$\,\cup\,$q+a & q$\,\cup\,$q+a \\
\midrule
Qwen3-VL-8B  & 39.9 & 38.4 & 51.2 & 41.2 \\
Qwen3-VL-30B & 34.8 & 37.4 & 46.9 & 40.6 \\
Qwen3.5-9B   & 25.2 & 32.5 & 40.5 & 34.2 \\
Qwen3.5-27B  & 44.5 & 40.5 & 53.7 & 41.7 \\
Gemma-3-12B  & 29.9 & 41.8 & 50.0 & 44.5 \\
Gemma-4-31B  & 44.5 & 42.8 & 55.4 & 45.3 \\
8B + GRPO    & 51.2 & 52.0 & 65.5 & 54.7 \\
\bottomrule
\end{tabular}
\caption{Oracle union of the two readings of the same model output, evidence recall at IoU
0.5 (n=719). The quotes and q+a columns repeat Table~\ref{tab:controls}; each union column
scores a question by the better of its two readings and diagnoses signal disjointness rather
than a system. The q$\,\cup\,$q+a column is the query-diversity control.}
\label{tab:union}
\end{table}

\paragraph{Resolver ablation and quote fidelity.}
Table~\ref{tab:resolver} resolves the same quotes with a token-F1 lexical matcher and with
Okapi BM25 over the parsed block texts, under the same whole-document scope and one-to-one
assignment as the production matcher, and lists the quote-fidelity rates. Exact means the
normalized quote is a substring of some single block text, which fails for quotes that span
block boundaries; fuzzy means a token-F1 of at least 0.8 against some block.

\begin{table}[t]
\centering
\footnotesize
\setlength{\tabcolsep}{2.8pt}
\begin{tabular}{lccccc}
\toprule
Backbone & exact & fuzzy & lex & BM25 & ours \\
\midrule
Qwen3-VL-8B  & 14.9 & 48.7 & 28.5 & 27.9 & 39.9 \\
Qwen3-VL-30B & 22.3 & 52.9 & 24.3 & 24.3 & 34.8 \\
Qwen3.5-9B   & 40.0 & 57.2 & 20.7 & 20.6 & 25.2 \\
Qwen3.5-27B  & 36.3 & 57.4 & 32.4 & 32.2 & 44.5 \\
Gemma-3-12B  & 16.2 & 39.4 & 25.5 & 24.6 & 29.9 \\
Gemma-4-31B  & 29.4 & 53.1 & 33.9 & 33.5 & 44.5 \\
8B + GRPO    & 20.6 & 50.6 & 37.6 & 36.8 & 51.2 \\
\bottomrule
\end{tabular}
\caption{Quote fidelity (percent of quotes with an exact or fuzzy block match) and evidence
recall at IoU 0.5 under lexical, BM25, and multimodal (ours) resolvers on identical quotes.}
\label{tab:resolver}
\end{table}

\section{Qualitative Examples}
\label{app:cases}
\begin{figure*}[htbp]
\centering
\includegraphics[width=\textwidth]{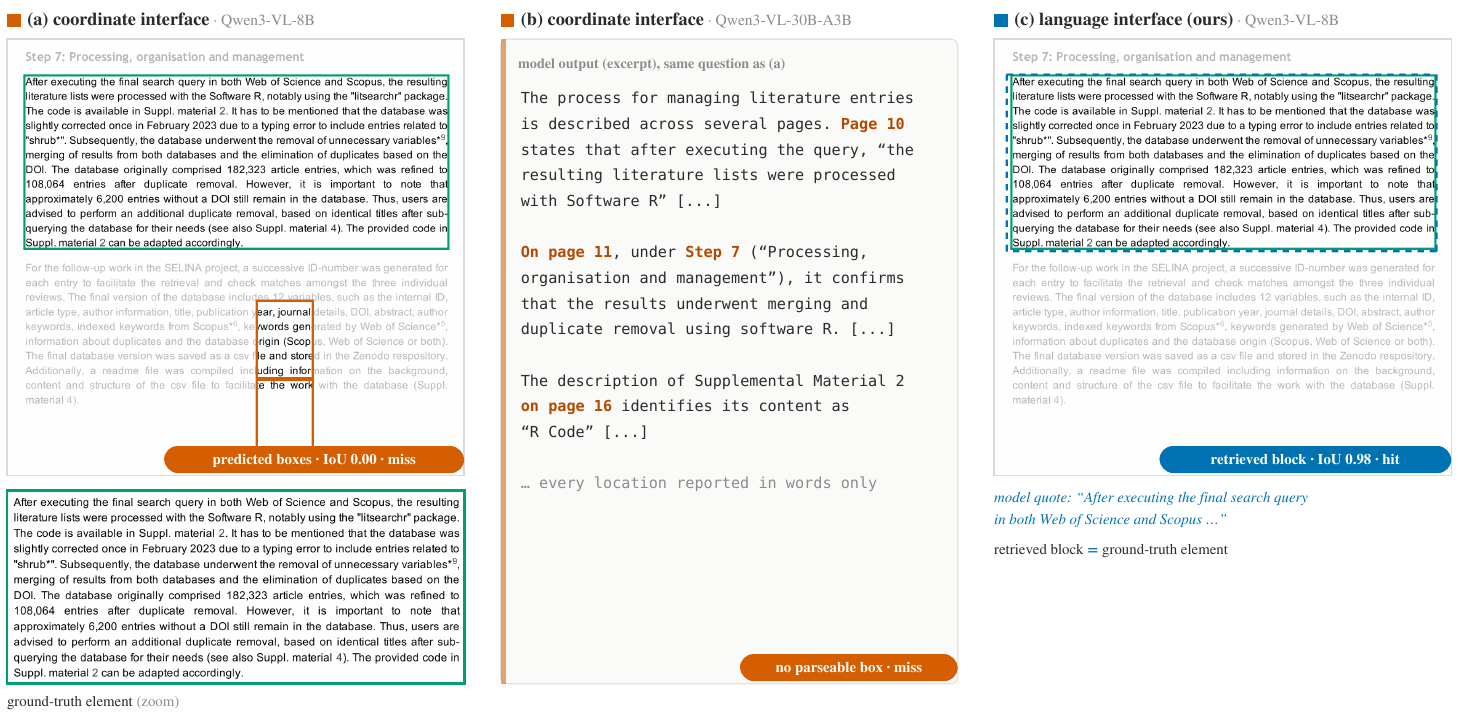}
\caption{Real model outputs for the same question on the same document (which programming tool
was used in the data-management phase). Dimmed page regions are shown only for context; the
clear regions are the annotated ground-truth element (green) and the model's citation. (a)
Qwen3-VL-8B under the coordinate interface answers correctly and reaches the correct page, but
emits two stacked template-like boxes below the evidence (IoU 0); the zoom shows the annotated
element, whose text contains the answer. (b) Qwen3-VL-30B-A3B, asked for coordinates, answers and describes every
evidence location in words, page numbers and a step heading, emitting no parseable box. (c)
Qwen3-VL-8B under the language interface quotes the evidence verbatim, and retrieval returns
the exact ground-truth block (IoU 0.98).}
\label{fig:cases}
\end{figure*}

Figure~\ref{fig:cases} shows real outputs for one document question, all three produced by the
models we evaluate rather than constructed for illustration. The three panels correspond to the
three behaviors the paper distinguishes: a coordinate answer that reaches the correct page but
places its box away from the evidence, a coordinate prompt answered with a verbal description
of the location instead of a box, and the same evidence quoted verbatim and resolved by
retrieval to the annotated block. Together they show that the information needed for
attribution is present in all three cases and that only the interface differs
(Section~\ref{sec:analysis}).

\section{Citation Precision and Residual Errors}
\label{app:residual}
This section supports the precision trade-off of Section~\ref{sec:results} and the
error analysis of Section~\ref{sec:training-fix}.

\paragraph{Scoring protocol.}
We inherit the matching rule, the metric definitions, and the judge prompts from CiteVQA, and
state here both what they compute and the two choices we add on top. All box metrics use the
same matching rule: a cited box matches an annotated element when the
two lie on the same page and their IoU is at least 0.5. Recall is computed per question as the
number of \emph{necessary} elements matched by at least one citation divided by the number of
necessary elements, then macro-averaged over questions. Precision replicates the released
scorer's loop verbatim: its numerator counts \emph{annotated elements that some citation hits},
and its denominator is the \emph{number of boxes the model cited}. The two therefore range over
different objects, and a single box that covers two annotated elements contributes two to the
numerator and one to the denominator, so the quantity is not a standard precision and can
exceed one in rare cases. Per-question F1 is the harmonic mean of these two quantities.

The two choices we add are the following, and we apply them identically to every condition. First, recall is scored against necessary evidence while
precision is scored against all annotated evidence, including the optional supporting elements,
so that citing genuinely supporting material the annotators marked optional is not counted as
an error. Second, we score citations against whichever page the model names, without correcting
page indices. If anything the first choice disadvantages the language interface, which cites
more regions per response and therefore has more opportunity to land outside the annotation.

\begin{figure}[htbp]
\centering
\includegraphics[width=\columnwidth]{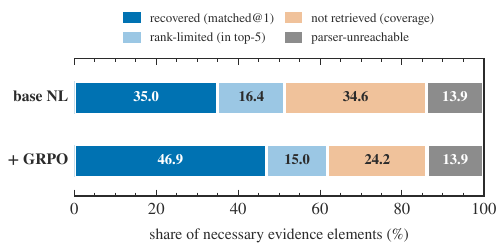}
\caption{Element-level decomposition of the 1{,}034 necessary evidence elements (IoU 0.5),
before and after GRPO. Recovered: matched by the current top-1 assignment. Rank-limited: the
correct block is in the retriever's top five but loses the assignment. Not retrieved: absent
from every quote's top five (a coverage failure). Parser-unreachable: no semantic block
overlaps the annotation. Training grows the recovered share mostly by shrinking the
not-retrieved bucket; the parser-unreachable share is constant by construction.}
\label{fig:residual}
\end{figure}

\paragraph{Citation precision and F1.}
Citation precision rises from between 0.3 and 8.6 under the coordinate
interface to between 22.9 and 52.7 under the language interface, with 1.0 to 3.4 citations per
response under coordinates and 1.1 to 3.3 under language. Per-question box F1 follows the same
ordering (0.2 to 6.7 under coordinates, 21.5 to 39.1 under language). The GRPO model cites 4.6
regions per response at 26.5 precision and 28.8 F1, against 2.1, 37.6, and 32.2 for its base
model, as discussed in Section~\ref{sec:results}. Precision is the counterweight that makes
recall interpretable: citing every parsed block would attain the ceiling recall of
Table~\ref{tab:threshold} by construction, at a precision below one percent. Truncating the
trained model's citations to the baseline's budget, whether by retrieval similarity or by the
model's own citation order, does not preserve its recall advantage, so the training gain
operates through broader coverage rather than through more precise individual citations,
consistent with the element-level decomposition of Figure~\ref{fig:residual} and with the
judge-based analysis of Section~\ref{sec:results}. Table~\ref{tab:budget} gives the
matched-budget numbers. Precision counts any citation outside a labeled element as wrong. On the 597 questions
where the trained model cites more regions than the baseline, judged relevance rises from 52.3
to 60.1 and the stricter SAA variant from 19.6 to 27.6; on the 119 questions where it does not
cite more, recall falls from 47.6 to 40.5.

\begin{table}[t]
\centering
\footnotesize
\setlength{\tabcolsep}{4pt}
\begin{tabular}{lccc}
\toprule
Citations scored & Rec$_{.5}$ & Prec & F1 \\
\midrule
Base model (2.1 per response) & 39.9 & 37.6 & 32.2 \\
GRPO, all (4.6 per response) & 51.2 & 26.4 & 28.7 \\
GRPO, first $k$ in output order & 30.6 & 33.7 & 24.9 \\
GRPO, top $k$ by similarity & 30.9 & 33.2 & 24.9 \\
\bottomrule
\end{tabular}
\caption{The trained model at the base model's citation budget (n=719); $k$ is the number of
citations the base model produced for the same question.}
\label{tab:budget}
\end{table}

\paragraph{Element-level residual decomposition.}
This analysis asks, for each necessary evidence element in the verified evaluation set, why
the language interface did or did not recover it. Every one of the 1{,}034 elements falls into
exactly one of four categories. \emph{Recovered} means the element is matched by the citation
the pipeline actually produced. \emph{Rank-limited} means the correct block does appear among
the retriever's top five candidates for some quote, but the one-to-one assignment awards that
block to a different quote, so the element is lost at the assignment step rather than at
retrieval. \emph{Not retrieved} means the correct block appears in no quote's top five at all,
which happens when the model does not quote that piece of evidence; this is a failure of
coverage rather than of ranking. \emph{Parser-unreachable} means no semantic block overlaps the
annotation at IoU 0.5, so no assignment could have recovered the element in the first place.

For the base model, 13.9 percent of elements are parser-unreachable, 48.5 percent are not
retrieved (a figure that includes the unreachable ones, since they are absent from every
candidate list as well), and 16.4 percent are rank-limited. Two readings follow. First, the
parser-unreachable share overstates the ceiling: 95.6 percent of those elements are fully
contained inside some block, so the failure is a granularity mismatch between the annotation
and the parser's units rather than a missing block, and a containment criterion in place of
IoU 0.5 would make them reachable. We nevertheless report IoU 0.5 as the primary criterion
because it is the published standard. Second, the residual is dominated by coverage rather
than by ranking, the not-retrieved share being about three times the rank-limited share.

Applying the same decomposition after GRPO isolates what training changed. Of the 11.9-point
gain in recovered elements, 10.4 points come out of the not-retrieved bucket, that is, from
evidence the base model never quoted and the trained model does. The rank-limited share barely
moves, and the parser-unreachable share is constant by construction. This is the behavior that
the coverage term of Eq.~\ref{eq:reward} rewards, and the training dynamics agree: the coverage
score and the number of citations per response rise the most over training
(Figure~\ref{fig:rewarddyn}).

\paragraph{Residual by question and evidence type.}
Table~\ref{tab:buckets} splits the decomposition of the base model by question type and by
evidence type. Figure evidence is the least recovered (9.9 percent, against 38.1 for text and
35.4 for tables): it is the least often quoted and the least often reachable through the
parser, a direct consequence of verbatim quoting. Complex Synthesis and Quantitative Reasoning
elements are recovered more often than Factual Retrieval and Multimodal Parsing ones.

\paragraph{Failure examples.}
Three representative failures of the base model illustrate the categories. (i) \emph{Evidence
that cannot be quoted.} Asked for the marker colors and shapes of two series in a chart, the
model writes a description of the chart as its quote; no block contains that text, and the
quote resolves to a heading on the previous page (Figure~\ref{fig:sidebyside}, Case 2).
(ii) \emph{A quote that spans two blocks.} Asked to compare two table entries on a slide, the
model concatenates the slide title with a table row; the quote resolves to the title block on
the correct page (IoU 0 with the table). (iii) \emph{Evidence that is never quoted.}
Asked for emission thresholds that define a major source, the model quotes three related
passages from pages 6 to 8, while the annotated paragraph is on page 17; the element is absent
from every quote's candidates and the answer is also wrong.

\begin{table}[t]
\centering
\footnotesize
\setlength{\tabcolsep}{2.0pt}
\begin{tabular}{lrcccc}
\toprule
 & $N$ & rec. & rank-lim. & not retr. & unreach. \\
\midrule
Complex Synthesis      & 532 & 38.0 & 17.1 & 33.1 & 11.8 \\
Multimodal Parsing     & 241 & 29.9 & 13.3 & 37.3 & 19.5 \\
Factual Retrieval      & 179 & 30.2 & 17.9 & 37.4 & 14.5 \\
Quant. Reasoning       & 82  & 43.9 & 15.9 & 30.5 & 9.8 \\
\midrule
text   & 685 & 38.1 & 15.9 & 32.0 & 14.0 \\
table  & 271 & 35.4 & 15.9 & 38.4 & 10.3 \\
figure & 71  & 9.9  & 18.3 & 46.5 & 25.4 \\
\bottomrule
\end{tabular}
\caption{Element-level residual decomposition of the Qwen3-VL-8B language run by question type
and by evidence type (percent of $N$ necessary elements): recovered, rank-limited, not
retrieved, and parser-unreachable, as defined above. Seven elements of rarer annotation types are omitted from the evidence-type rows.}
\label{tab:buckets}
\end{table}

\section{Selective Prediction on Citation Similarity}
\label{app:abstain}
\begin{figure}[htbp]
\centering
\includegraphics[width=\columnwidth]{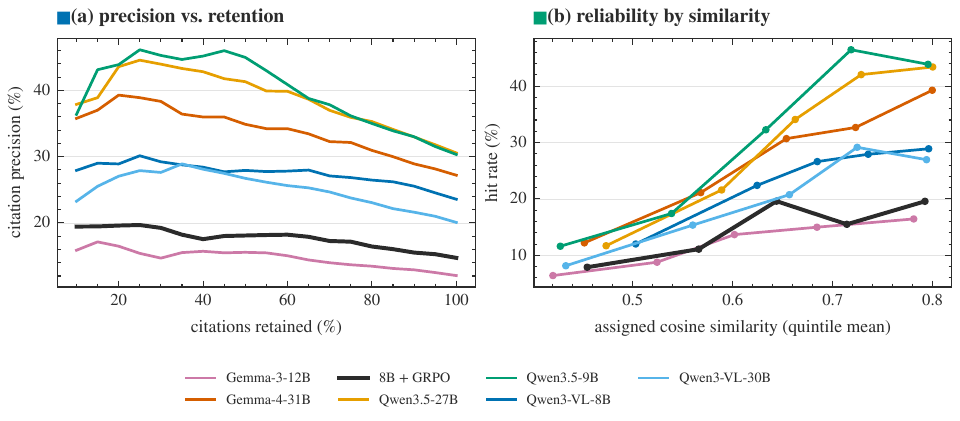}
\caption{Judge-free selective prediction over assigned citations. (a) Citation precision
against necessary evidence as low-similarity citations are abstained. (b) Hit rate by
similarity quintile; the monotone trend shows the similarity is informative about citation
correctness.}
\label{fig:abstention}
\end{figure}

The assignment of Eq.~\ref{eq:assign} returns a block for every quote, including quotes that
match nothing in the document. The assigned cosine similarity provides a usable abstention
signal: citation hit rates rise monotonically with similarity for every backbone (from roughly
10 percent in the lowest quintile to between 16 and 47 percent in the highest), and retaining
only the most similar citations raises citation precision, for example from 30.6 to 43.6
percent for Qwen3.5-27B when keeping the top fifth (Figure~\ref{fig:abstention}). A deployed
system can therefore trade coverage for reliability by abstaining below a similarity threshold
instead of always presenting a citation.

\section{ANLS as an Answer Metric}
\label{app:anls}
Table~\ref{tab:anls} reports ANLS for all 13 conditions, computed between the model answer
and the gold answer. The metric shows a floor effect: CiteVQA gold answers are paragraph-length
and abstractive (median 264 characters in English and 93 in Chinese), so generated answers
rarely share enough characters with them to pass the 0.5 threshold. ANLS stays between 0.0
and 8.7 while the judged answer scores range from 37 to 92, and the per-question correlation
between NLS and the judge stays at or below 0.34. The main results therefore use the CiteVQA
judge protocol.

\begin{table}[t]
\centering
\footnotesize
\setlength{\tabcolsep}{3pt}
\begin{tabular}{lcccc}
\toprule
Condition & ANLS & soft NLS & $r$ & Ans \\
\midrule
Qwen3-VL-8B coord.  & 0.1 & 14.3 & 0.23 & 59.6 \\
Qwen3-VL-30B coord. & 0.5 & 14.9 & 0.13 & 61.9 \\
Qwen3.5-9B coord.   & 0.0 & 6.4  & 0.24 & 89.7 \\
Qwen3.5-27B coord.  & 0.0 & 6.0  & 0.24 & 91.8 \\
Gemma-3-12B coord.  & 1.3 & 18.0 & 0.33 & 39.5 \\
Gemma-4-31B coord.  & 4.5 & 25.4 & 0.29 & 78.7 \\
\midrule
Qwen3-VL-8B lang.   & 7.0 & 27.0 & 0.33 & 53.4 \\
Qwen3-VL-30B lang.  & 6.8 & 21.5 & 0.23 & 60.3 \\
Qwen3.5-9B lang.    & 8.7 & 20.0 & 0.11 & 89.6 \\
Qwen3.5-27B lang.   & 7.9 & 24.8 & 0.24 & 90.0 \\
Gemma-3-12B lang.   & 1.9 & 18.4 & 0.34 & 37.2 \\
Gemma-4-31B lang.   & 7.4 & 26.6 & 0.24 & 77.0 \\
8B + GRPO lang.     & 7.6 & 27.3 & 0.31 & 60.4 \\
\bottomrule
\end{tabular}
\caption{ANLS and its thresholdless variant (soft NLS) on the judged subset, with the Pearson
correlation $r$ between per-question NLS and the judge score, and the judged answer score (Ans)
for reference.}
\label{tab:anls}
\end{table}

\clearpage
\onecolumn
\input{appendix_prompts}

\end{document}

%% file: appendix_prompts.tex
\section{Prompts}
\label{app:prompts}
This appendix reproduces, verbatim, every prompt used in the paper.

\subsection{Coordinate-interface system prompt}
\label{app:prompt-coord}
Taken verbatim from the official CiteVQA release (\texttt{prompts/infer\_system.txt}), including its element-level instructions and in-context examples.
\begin{promptbox}
# Document Analysis Assistant

Answer the question based on the provided PDF page images, and cite the evidence regions in your answer.

## Evidence Citation Rules

1. Evidence must be at the **element level**: a complete paragraph, a complete table, a complete image, or a complete note. Do not select partial text from a paragraph or a single row from a table, and do not select an entire page or spanning multiple tables/paragraphs. Note: This is very important and will directly affect your score.
2. For **tables and images**, if there are captions or footnotes, they need to be annotated as **separate evidence** with their own bbox, not merged into the table/image bbox.
3. Each piece of cited evidence text should be followed by a `<bbox />` tag indicating the evidence location.
4. When an inference step relies on multiple pieces of evidence, use multiple `<bbox />` tags separately.
5. Pure reasoning/calculation steps do not need `<bbox />`.

## Annotation Format

```
<bbox page="page_number" x1="left" y1="top" x2="right" y2="bottom" />
```

Page numbers start from 1 (note: ignore original page numbers); coordinates are relative coordinates on the page image, range 0-1000.

## Examples

**Question:** What is the net change in the company's precision copper tube production capacity from 2021 to 2024?

**Answer:**

According to the main text, the company's precision copper tube production capacity increased from 798,000 tons in 2021 to 1.31 million tons in 2024:
<bbox page="1" x1="536" y1="65" x2="642" y2="656" />

Therefore, the net change = 1.31 - 0.798 = 0.512 million tons.

Additionally, according to "Table 1: Production line renovation will reduce the company's costs", the per-ton comprehensive cost is expected to decrease by 700 yuan/ton after the production line renovation:
<bbox page="8" x1="584" y1="65" x2="598" y2="371" />
<bbox page="8" x1="598" y1="59" x2="712" y2="670" />

## Final Reminder
Evidence must be a complete paragraph, a complete table, a complete image, or a complete note. Do not select partial rows from a paragraph or a single row from a table, and do not select an entire page or spanning multiple tables/paragraphs. Note: This is very important and will directly affect your score.
\end{promptbox}

\subsection{Language-interface system prompt (ours)}
\label{app:prompt-nl}
The GRPO rollouts use the same instruction (identical wording, different line wrapping in the training copy).
\begin{promptbox}
# Document Analysis Assistant

Answer the question based on the provided PDF page images. For EVERY piece of evidence that supports your answer, QUOTE the exact text span from the document (verbatim, as it appears), so it can be located in the source.

## Rules
1. Each evidence quote must be a verbatim text span copied from the page (a sentence, a table row/cell, a caption, or a note). Do not paraphrase.
2. Quote the evidence at the element level: enough text to identify the region, not a single word and not a whole page.
3. If multiple regions support the answer, give multiple quotes.

## Output Format
Return a single JSON object exactly in this form:
{"answer": "<your full answer text>", "evidence": [{"page": p, "quote": "<verbatim text>"}, ...]}
where p is the 0-based page index in the order shown.
\end{promptbox}

\subsection{Evaluation judge prompt: answer accuracy}
\label{app:prompt-judge-qa}
Official CiteVQA judge prompt, given to Gemini-3.5-Flash. The system prompt is followed by the user-message template.
\begin{promptbox}
## Task
You are a multimodal QA evaluation expert. Your task is to evaluate the overall quality of the answer. Provide your evaluation in the form of "reasoning" and "score". Evaluation should be based solely on the standard answer, without introducing your own external knowledge.
You will receive a question, a standard answer, and the model's generated answer.

## Evaluation Criteria
**BE STRICT. Most answers are not as good as they appear. When in doubt, choose the lower score.**
- 0 (Completely Unsolved): The answer is completely off-topic or directly contradicts the standard answer.
- 1 (Mostly Unsolved): The answer has extremely low relevance, providing almost no valuable information.
- 2 (Partially Solved): The answer covers some aspects but misses key information or has obvious factual errors. **Many "okay" answers fall here — do not over-rate.**
- 3 (Acceptable): The answer covers the core facts but is incomplete, lacks necessary details, or has minor errors. **Only give this when the answer is genuinely useful despite clear gaps.**
- 4 (Good): The answer clearly covers all key points with rigorous logic. Near-complete and accurate. **Reserve for strong answers. Do not hand out freely.**
- 5 (Excellent): Complete, accurate, and perfectly structured and the answer must not be significantly more verbose than the standard answer.  **Extremely Difficult to reach. Do not give 5 unless truly prefect in every dimension.**

## Important Notes
- Ignore phrases like "cited from" or "from" that may appear in the model's generated answer — they are irrelevant.
- **DO NOT penalize the answer based on the language it is written in.** Chinese, English, or mixed — score the content only.
- Only the exact facts in the standard answer count. Extra details beyond the standard answer do NOT improve the score.

## Output Format    
Please output two lines for the results: the first line is your reasoning for the score, and the second line is the score. Strictly follow this format without any additional content.

# Output Example  
A reason why you choose this score (from 0 to 5).
```<qa_acc>X</qa_acc>```

----- user message template -----
Question: {question}

Standard Answer: {standard_answer}

Model's Answer: {model_answer_no_bbox}
\end{promptbox}

\subsection{Evaluation judge prompt: evidence relevance}
\label{app:prompt-judge-rel}
Official CiteVQA judge prompt; the retrieved evidence crops are attached to the user message as images.
\begin{promptbox}
## Task
You are a professional DocVQA quality evaluation expert. Your task is to evaluate whether the PDF screenshots referenced in the answer can effectively support the corresponding answer content. You need to determine whether the visual information (text, charts, data) in the screenshots is consistent with the facts mentioned in the answer.

You will receive a question, a standard answer (without images), and the model's generated answer with interleaved images.

## Evaluation Dimensions
- Truthfulness: Does the screenshot contain the key data or descriptions mentioned in the answer?
- Sufficiency: Does the screenshot provide sufficient evidence for the conclusion, or is it taken out of context?
- Localization Accuracy: Does the screenshot precisely cover the answer source, or does it contain irrelevant information?
- Alignment: Does the screenshot exactly match the text being cited? Any misalignment is a flaw.

## Scoring Criteria
**BE STRICT. A score of 5 is extremely rare and requires perfection. Most good answers should score 3-4.**
- 0: No support at all. The screenshot content is completely irrelevant to the answer.
- 1: Extremely weak support. The screenshot only mentions vague related concepts without specific data.
- 2: Weak support. The screenshot contains partial key data, or has significant quality issues.
- 3: Moderate support. The screenshot covers most of the evidence but has flaws (e.g., includes irrelevant content, slight misalignment with cited text, or captures too much/too little).
- 4: Good support. The screenshot contains the core evidence with minor flaws. This is where most correct answers should score.
- 5: **PERFECT support (extremely rare)**. The screenshot must be **flawless**: precise bounding box that exactly covers the cited text, no extra content, no skewing, no misalignment, and the evidence perfectly matches what is claimed. **Only give 5 when every single detail is perfect.**

## Important Notes
- Be conservative with scores. If you hesitate between two scores, choose the lower one.
- A slightly off-center crop, a small amount of extra content, or minor misalignment = score 3-4, NOT 5.
- Score 5 should only be given when the bounding box is pixel-perfect and the evidence is exactly what was cited.

## Output Format    
Please output two lines for the results: the first line is your reasoning for the score, and the second line is the score. Strictly follow this format without any additional content.

# Output Example  
A reason why you choose this score (from 0 to 5).
```<relevance_score>X</relevance_score>```

----- user message template -----
Question: {question}

Standard Answer (without images, no need to evaluate): {standard_answer}

Model's Answer with Images: {answer_with_images}
\end{promptbox}

\subsection{Training reward judge prompt}
\label{app:prompt-reward}
Scores $a$, $e_{\mathrm{rel}}$, and $e_{\mathrm{cov}}$ of Eq.~\ref{eq:reward}. The judge sees the question, the gold answer, the model answer, and the retrieved crops; it never sees the model's raw quotes.
\begin{promptbox}
You are a strict evaluator for a document QA system. The system answered a question about a document and cited EVIDENCE regions, shown to you as cropped images taken from the document pages (these are the regions the system localized). You are also given the REFERENCE (ground-truth) answer.

Score these THREE aspects, each an integer 0-5. Use the FULL range; do not default to 5.

1. answer_correctness: does the SYSTEM ANSWER semantically match the REFERENCE answer? 5=equivalent, 3=partially correct or missing detail, 0=wrong/unrelated. Ignore wording, order, formatting; judge meaning.

2. evidence_relevance: are the cited crops the correct SOURCE that supports the reference answer? 5=a crop clearly contains the exact fact/figure/table-cell/text the reference answer comes from, 3=crops are on-topic but do not pinpoint it (related context only), 0=crops are irrelevant, or no evidence was provided.

3. evidence_coverage: do the cited crops TOGETHER cover ALL the information the reference answer requires? For a multi-part answer (several values/items/conditions), 5=every needed piece appears in some crop, 3=about half present, 1=only a small part, 0=key pieces missing or no evidence. For a single-fact answer, 5 if the one needed region is present, else 0.

Output ONLY a JSON object: {"answer_correctness":<0-5>,"evidence_relevance":<0-5>,"evidence_coverage":<0-5>,"reason":"<one short sentence>"}
\end{promptbox}